\documentclass[a4paper,fleqn]{cas-sc}

\usepackage[authoryear]{natbib}
\usepackage{color, soul}
\usepackage{multirow}
\usepackage{geometry}
\usepackage{pdflscape}
\usepackage{float}%

\def\tsc#1{\csdef{#1}{\textsc{\lowercase{#1}}\xspace}}
\tsc{WGM}
\tsc{QE}

\usepackage{verbatim}

\newcolumntype{P}[1]{>{\centering\arraybackslash}p{#1}}
\begin{document}
\let\WriteBookmarks\relax
\def\floatpagepagefraction{1}
\def\textpagefraction{.001}

\shorttitle{Benchmarks for Retrospective ADS Crash Rate Analysis}    

\shortauthors{Scanlon et al.}  

\title [mode = title]{
  Benchmarks for Retrospective Automated Driving System Crash Rate Analysis Using Police-Reported Crash Data
}

\author[1]{John M. Scanlon}[
]
\ead{johnscanlon@waymo.com}
\cormark[1]
\credit{Conceptualization, Data curation, Writing - original draft}

\author[1]{Kristofer D. Kusano}[
  orcid=0000-0003-4976-6114
]
\ead{kriskusano@waymo.com}
\credit{Conceptualization, Data curation, Validation, Writing - review \& editing}

\author[1]{Laura A. Fraade‐Blanar}[
  orcid=0000-0002-7211-8206
]
\ead{lblanar@waymo.com}
\credit{Writing - original draft, Writing - review \& editing}

\author[1]{Timothy L. McMurry}[
  orcid=0000-0001-5912-5092
]
\ead{tmcmurry@waymo.com}
\credit{Data curation, Formal Analysis, Methodology, Validation}

\author[1]{Yin-Hsiu Chen}[]
\ead{yinhsiuchen@waymo.com}
\credit{Data curation, Methodology, Validation}

\author[1]{Trent Victor}[
  orcid=0000-0002-9550-2411
]
\ead{trentvictor@waymo.com}
\credit{Conceptualization, Supervision, Writing - review \& editing}

\affiliation[1]{organization={Waymo, LLC.},
            addressline={1600 Ampatheater Parkway}, 
            city={Mountain View},
            postcode={94043}, 
            state={CA},
            country={USA}}

\cortext[1]{Corresponding author}

\begin{abstract}
Objectives: With fully automated driving systems (ADS; SAE level 4) ride-hailing services expanding in the U.S., we are now approaching an inflection point in the history of vehicle safety assessment. The process of retrospectively evaluating ADS safety impact (as seen with seatbelts, airbags, electronic stability control, etc.) can start to yield statistically credible conclusions. An ADS safety impact measurement requires a comparison to a “benchmark” crash rate. Most benchmarks generated to-date have focused on the current human-driven fleet, which enable researchers to understand the impact of the introduced ADS technology on the current crash record status quo. This study aims to address, update, and extend the existing literature by leveraging police-reported crashes to generate human crash rates for multiple geographic areas with current ADS deployments. 

Methods: All of the data leveraged is publicly accessible, and the benchmark determination methodology is intended to be repeatable and transparent. Generating a benchmark that is comparable to ADS crash data is associated with certain challenges, including data selection, handling underreporting and reporting thresholds, identifying the population of drivers and vehicles to compare against, choosing an appropriate severity level to assess, and matching crash and mileage exposure data. 

Results: Consequently, we identify essential steps when generating benchmarks, and present our analyses amongst a backdrop of existing ADS benchmark literature. One analysis presented is the usage of established underreporting correction methodology to publicly available human driver police-reported data to improve comparability to publicly available ADS crash data. We also identified several important crash rate dependencies (geographic region, road type, and vehicle type), and show how failing to account for these features in ADS comparisons can bias results. 

Conclusions: Working with police-reported crash data to create crash rate benchmarks is fraught with challenges. Researchers should be cautious in their selection of crash rate benchmarks. We present these challenges, discuss their consequences, and provide analytical guidance for addressing them. This body of work aims to contribute to the ability of the community - researchers, regulators, industry, and experts - to reach consensus on how to estimate accurate benchmarks.

\end{abstract}


\maketitle

\section*{INTRODUCTION}\label{section:intro}

\subsection*{The Burden of Motor Vehicle Crashes}\label{subsection:burden}

Automotive crashes are a crisis on U.S. public roadways \citep{usdot2022nrss}. In 2022, the U.S. experienced 5.9 million police-reported crashes \citep{ncsa2023a} and 42,514 fatalities \citep{ncsa2023b}. Several vehicle solutions, including improved vehicle crashworthiness and advanced driver assistance systems (ADAS), have all contributed toward reducing this motor vehicle crash burden \citep{glassbrenner2009lives, fildes2015effectiveness, lie2006effectiveness, teoh2011iihs, strandroth2012effects, cicchino2017effectiveness, cicchino2018effects, isaksson2015evaluation}. Despite the adoption and improvement of passive and active safety systems, the U.S. had the most traffic fatalities in 2021 since 2005 \citep{stewart2023overview}. 

\subsection*{Safety Impact of Automated Driving Systems}

Automated driving systems (ADS), specifically SAE level 4 ADS used in fully automated ride-hailing services without a human behind the steering wheel, have been deployed in multiple urban metropolitan areas throughout the U.S. In an SAE level 4 ADS, when the system is active, the ADS can handle the entire dynamic driving task, and a human driver is not required to continuously monitor the driving task or urgently respond \citep{SAE_J3016}. While an ADS is in the development phase, prospective and design-based methods are used to evaluate the safety of the system before widespread deployment \citep{favaro2023building, webb2020waymo}. Scaled, rider-only (RO) deployments - without a human in the driver seat - enable a new phase of ADS evaluation: retrospective safety impact. At this deployment stage, the performance of the system can be evaluated without the confounding effect of a human available to take control. 

There is a long history in the automotive safety literature of retrospective safety impact analysis, including the assessment of airbags, seatbelts, electronic stability control, and automated emergency braking \citep{glassbrenner2009lives, fildes2015effectiveness, lie2006effectiveness, teoh2011iihs, strandroth2012effects, cicchino2017effectiveness, cicchino2018effects, isaksson2015evaluation}. In retrospective safety impact analyses, crash and injury rates are evaluated against some benchmark rate(s). A common point of comparison is the relative performance of the introduced technology to the current rate of crashes and injuries. This evaluation enables researchers to make statistical claims regarding the impact of the introduction of the technology relative to the driving population at-large. 

To execute retrospective safety impact on ADS technology, enough driving exposure must be accumulated to where there is some deviation from an expected crash and/or injury rate that can be detected by statistical tests \citep{kalra2016driving, lindman2017basic}. Because of this mileage dependency, we are in the early stages of retrospective ADS evaluation \citep{victor2023safety, dilillo2023comparative}. A valid comparison between ADS and benchmark is enabled by consistent safety-relevant reporting standards, accounting for influential factors, and credible statistical testing. 

\subsection*{Benchmarks for ADS Performance Evaluation}

An ideal safety impact analysis for an ADS would compare crash rates of an ADS operated fleet with a benchmark crash rate of the existing fleet (human-driven vehicles). The goal of the benchmark is to create a fair and accurate representation of that existing fleet within the context of the ADS operational design domain (ODD) conditions. Two main dimensions should be considered when generating a benchmark with which to compare an ADS to: (1) what data can be relied upon for generating benchmarks? and (2) what minimum methodological requirements are required to generate a valid, apples-to-apples comparison? 

Benchmark creation efforts - for both ADS and ADAS - have largely considered three primary data sources: Insurance, naturalistic driving study, and police-reported. Appendix \ref{appendix:crash_data_sources} discusses each of these data sources, including their scale, composition, and availability. This current study examined the utility of police-reported crash databases and aggregated mileage estimates from government agencies for generating benchmarks. These data sources have the distinct advantage of being publicly accessible for wide usage, representative of a large amount of driving miles, and inclusive of many geographic domains where ADS-equipped vehicles are currently deployed. 

Several studies have generated benchmarks for ADS evaluation using police-reported crash and publicly-accessible mileage data. A summary of these studies can be found in Appendix \ref{appendix:prev_benchmark_studies}. The first contribution of this current study is the methodological handling and presentation of these key challenges that researchers face when leveraging this publicly accessible data, which can ultimately unreasonably bias results if not properly handled. The opportunity for ADS comparison to a police-reported crash benchmark does bring with it certain challenges that include: 

\begin{enumerate}
  \item Analytical error of comparing crash-level and vehicle-level rates,
  \item Matching reporting thresholds,
  \item Accounting for underreporting,
  \item ODD-matching, and
  \item Mismatched crash and mileage data.
\end{enumerate}

A second contribution of the current study is that it supplements this existing body of research using the most recent police-reported benchmarks and within multiple geographic regions where ADS technology is currently deployed. Each of the identified potential biases was addressed within the analysis and the quantitative impact of accounting for them was demonstrated. We specifically examined the opportunity to generate lower-severity-inclusive crash benchmarks from these police-reported data, such as “any property damage, injury, or fatality” by correcting for underreporting. These lower-severity benchmarks are important because they represent the reporting threshold for most ADS crash data (i.e., any property damage with no lower reporting threshold). We also established rates for subsets of police-reported crashes of higher severity outcomes. 

\subsection*{Objectives and Research Questions}

The objective of this paper was to leverage police-reported data to create crash rate benchmarks for evaluating ADS performance against the current human crash rate status quo. Several research questions were posed. First, what are the effects of methodological choices made in generating benchmark rates? The second research question is: how can police-reported data be used to generate a benchmark that is comparable to ADS crash data, which has no lower property damage limit for reporting? Third, what are crash rate benchmarks for evaluating ADS performance within the Phoenix, San Francisco, and Los Angeles driving environments (three active deployment locations for ADS technology)? 

\section*{METHODS}\label{section:methods}

\subsection*{Data Sources}

All crash and mileage data used in the current study are publicly accessible. Crash data - at a minimum - contains crash, vehicle, and person data that can be combined to answer various research questions. For crash data, this study followed the general best practice of downloading and processing the published raw data tables. This study focused exclusively on data collected from 2022. This year of data was the most up-to-date year available in all data sources.

\subsubsection*{National data}

In the United States, police-reported databases are generally managed by federal, state, and local agencies. Nationally, two crash database sources were leveraged to generate crash rates. First, the Crash Report Sampling System (CRSS) was used to generate non-fatal, police reported crash counts \citep{ncsa2023a}. The CRSS database is annually compiled by NHTSA to quantify the total number of police-reported crashes on US roadways. According to NHTSA: ``By restricting attention to police-reported crashes, CRSS concentrates on those crashes of greatest concern to the highway safety community and the general public.'' The database is compiled by taking a probability sample of crashes within multiple jurisdictions, and each crash is then assigned a weight to enable nationally-representative estimates to be made. Second, although CRSS contains (sampled) fatal crashes, NHTSA also compiles the Fatality Analysis Reporting System (FARS), which contains a census of fatal motor vehicle crashes on US public roadways \citep{ncsa2023b}. To be included in FARS, at least one person must have been fatally injured. Both CRSS and FARS have unique data elements, but are generally similar in structure and formatting, which helps facilitate their combination into a single, nationally representative crash compilation. Both CRSS and FARS are publicly accessible for download \citep{nhtsa2023a}. 
For a national estimate of vehicle miles traveled (VMT), the U.S.\ Department of Transportation (DOT) Federal Highway Administration (FHWA) highway statistics series was used \citep{fhwa2023b}. This annually compiled publication contains multiple tabulations of various highway data, including VMT data compiled as part of the Highway Performance Monitoring System (HPMS). The data, itself, is compiled, processed, and verified by a cooperation of local, state, and federal agencies. From this series, the ``VM-2'' and ``VM-4'' tables were used. When combined, these two tables provide state-level VMT by functional system, vehicle type, and population density. 

\subsubsection*{State data}

Three counties where ADS technologies are currently deployed were considered for this analysis: Maricopa, Arizona, San Francisco, California, and Los Angeles, California. The Arizona Department of Transportation (ADOT) annually compiles a census of all police-reported crashes within the state that can be accessed via a public record request \citep{arizona2023a, arizona2023c}. In California, the Statewide Integrated Traffic Records System (SWITRS) database is publicly accessible for download via the California Highway Patrol’s (CHP) online portal (CHP, 2023). The SWITRS dataset is intended to be a full compilation of crashes that resulted in injury, where SWITRS ``processes all reported fatal and injury crashes that occurred on California’s state highways and all other roadways, excluding private property'' \citep{chp2021}. Police-reported crashes that are property damage only (PDO) are not fully accounted for in SWITRS due to the fact that ``some agencies report only partial numbers of their PDO crashes, or none at all'' \citep{chp2021}. Under California Vehicle Code \S 20008, injury crashes must be reported to CHP, and there is no requirement that PDO crashes be reported. This underreporting feature in California data is discussed further in the paper, but it is not clear whether Los Angeles and San Francisco fully report all of their PDO crashes to SWITRS.

For VMT estimates, three data sources were used. In Arizona, ADOT maintains the certified public miles (CPM) for roadways statewide \citep{arizona2023b}. In California, the California Department of Transportation (Caltrans) maintains Public Road Data (PRD) \citep{cadot2023}. The annually compiled VMT data from both Arizona’s CPM and California’s PRD are relied upon for each state’s HPMS submission to FHWA. Using the state mileage data sources rather than the FHWA mileage allows for estimating VMT at the county-level, which is not reported in the FHWA highway statistics data. Lastly, because the state mileage data tables do not specifically report the number of miles traveled by vehicle type, the FHWA VM-4 tables were used to estimate the proportion of the total driving miles attributable to passenger vehicles. The VM-4 tables provided an estimate of the relative proportion of driving mileage attributable to passenger vehicles in urban areas of California and Arizona. 

\subsection*{Data Processing}

\subsubsection*{Driver datasets}

Shown in Table~\ref{tab:crash_and_mileage_sources}, four main driver datasets of matched mileage-crashes were generated to estimate crashes and miles traveled in order to answer this study’s research questions. Aggregate statistics were then generated from each of these combined datasets. Because FARS is a census of all fatal crashes, CRSS was not needed to generate national fatal crash rates, and FARS was directly merged with FHWA mileage estimates. 

\begin{table*}[width=0.70\textwidth, cols=4]
  \centering
    \caption{Four unique combinations of crash and mileage data were leveraged to generate this study's benchmarks.}
    \begin{tabular*}{\tblwidth}{llll}
    \toprule
    \textbf{Geographic Region} & \textbf{Severity Level} & \textbf{Crash Source} & \textbf{Mileage Source} \\
    \midrule
    \multirow{2}{*}{National} & All Police-Reported & CRSS/FARS & FHWA   \\
    &  Fatal Only & FARS & FHWA  \\ 
    Maricopa County & All Police-Reported & ADOT & CPM + FHWA \\
    San Francisco County & All Police-Reported  & SWITRS & PRD+FHWA \\
    Los Angeles County & All Police-Reported  & SWITRS & PRD+FHWA \\
    \bottomrule
    \end{tabular*}
    \label{tab:crash_and_mileage_sources}
\end{table*}

\subsubsection*{ADS comparable drivers}

To isolate a comparable driving population to active, commercially-available ADS deployments, the current study considered two dimensions, road type and vehicle type, when selecting mileage and crash data. Each crash and mileage data source has its own unique set of data features (e.g., roadway definition, vehicle type definitions) for properly considering these two dimensions. When doing this analysis, the data should be subset in a way that (a) is comparable (calibrated) to the ADS deployment and (b) allows matching between the mileage and crash data. Appendix \ref{appendix:comp_pop_of_drivers} contains all subsetting procedures, including the specific data elements relied upon. 

For road type, although testing extends to higher speed roads, commercially-available Rider Only ADS technologies in the US currently only operate on ``surface streets'', where speed limits are generally lower. Accordingly, highways and interstates were identified and excluded in the mileage and crash data. 

For vehicle type, currently, commercially-available, ADS deployments consist exclusively of light-duty passenger vehicles, so accordingly, cars and light trucks / vans (LTV) were identified within the available data (e.g., heavy vehicles, low-speed vehicles, and motorcycles were excluded). These passenger vehicles were also identified for inclusion in both the mileage and crash data. 

\subsubsection*{Crash severity level}

Rates for several different crash severity levels were generated. We independently examined crashed vehicle rates of: 

\begin{enumerate}
	\item \textit{any property damage or injury}: an involved crash with “any property damage, injury, or fatality,” 
	\item \textit{police-reported}: a police report was filed for the involved crash,
	\item \textit{any-injury-reported}: any involved person sustained some level of reported injury,
	\item \textit{tow-away}: any of the involved vehicles was towed from the scene, 
	\item \textit{any airbag deployment}: any of the involved vehicles had an airbag deployment,
	\item \textit{suspected serious injury+}: any involved person had a suspected serious injury or was fatally injured, and 
	\item \textit{fatal}: any involved person was fatally injured. 
\end{enumerate}

Naturally, police-reported crashes were identified by their inclusion within the CRSS, FARS, ADOT, and California SWITRS crash databases. As noted previously, the CHP annual SWITRS report explicitly notes that, for certain agencies, the police-reported PDO crashes may not be fully accounted for, which would lead to an underestimation of associated police-reported rates for potentially affected areas (Los Angeles and San Francisco County). This limitation is discussed later in the paper. For both national estimates and Maricopa county, the police reported numbers presented are expected to reflect the true volume of crashes that resulted in a police-report. 

This study generated a benchmark crash rate of \textit{any property damage or injury} for comparison to ADS crash data. The NHTSA Standing General Order (SGO) 2021-01 requires ADS developers to report ``any physical impact between a vehicle and another road user (vehicle, pedestrian, cyclist, etc.) or property that results or allegedly results in any property damage, injury, or fatality'' \citep{nhtsa2023b}. This reporting requirement applies to both crashes involving the ADS, and crashes where the ADS ``contributes or is alleged to contribute (by steering, braking, acceleration, or other operational performance) to another vehicle’s physical impact with another road user or property involved in that crash.'' The latter cannot be readily determined from police-reported databases because non-damaged vehicles are not recorded in most crash databases, so this study focused only on generating a benchmark crash rate for only vehicles that sustained damage in a crash, and not vehicles that contributed to the crash but were not damaged. 

Determining \textit{any property damage or injury} rates from police-reported data requires a correction to account for underreporting and reporting thresholds. To generate this \textit{any property damage or injury} rate, we used two independent adjustment methods. First, we leveraged NHTSA's underreporting estimates from \citet{blincoe2015economic, blincoe2023economic} that were created using a combination of telephone surveyed persons \citep{mdavisandco}, insurance records, police reported crashes, and several other sources. \citet{blincoe2023economic} found that approximately 60\% of PDO crashes and 32\% of non-fatal injury crashes were not reported to police. We will refer to this adjustment as the ``Blincoe-adjusted'' estimate. Second, \citet{blanco2016automated} leveraged the SHRP-2 NDS study to generate estimates of overall underreporting of police-reportable (above a US \$1500 threshold) crashes without an associated police report. The motivation was that the observed NDS crash data could be more accurate than relying on surveyed persons for generating an underreporting correction, and also be more accurate than insurance data, which has some risk of underreporting of low amounts of property damage. In their analysis, \citet{blanco2016automated} estimated that 84\% of police-reportable PDO crashes were not reported by police. For this ``Blanco-adjusted'' estimate, we used 84\% underreporting of PDO crashes, and the \citet{blincoe2023economic} 32\% underreporting estimate for non-fatal injury crashes. Fatal crashes did not receive an adjustment factor when generating either the ``Blincoe-adjusted'' or the ``Blanco-adjusted'' rates, as it is assumed that all fatal crashes were reported. A complete listing of the associated adjustment factors can be found in Appendix \ref{appendix:underreport_adjust_fact}. 

\textit{Any-injury-reported}, \textit{tow-away}, \textit{any airbag deployment}, \textit{suspected serious injury+}, and \textit{fatal} crashes were identified by subsetting the police-reported data. The data selection routines used for each dataset are described in Appendix~\ref{appendix:severity_level}. We presented both \textit{unadjusted any-injury-reported} and \textit{Blincoe-adjusted any-injury-reported} crash rates, the latter of which applied the non-fatal injury crash underreporting weighting adjustment by \citet{blincoe2023economic} to the non-fatal proportion of the crashes.  No underreporting adjustment was applied to generate the \textit{tow-away}, \textit{suspected serious injury+}, or \textit{fatal} crash rates. 

\subsection*{Statistical Power Analysis}

To investigate how many miles would need to be driven to have a high probability of showing retrospective safety impact at a statistically significant level, we performed a power analysis. For this calculation, we considered a fictive ADS which has a known crashed vehicle rate that is some percentage of the human benchmarks derived in this study.  Using this setup, we calculated the mileage required to detect a statistically significant difference in performance.  We used sample size formulas based on the asymptotic normal approximation to the Poisson distribution and targeted 80\% statistical power using 2-sided tests with $\alpha = 0.05$.  For this exercise, we used the national benchmark crashed vehicle rates for \textit{Blanco-adjusted any property damage or injury}, \textit{police-reported}, \textit{Blincoe-adjusted any-injury-reported}, \textit{suspected serious injury+}, and \textit{fatal}, which cover the spectrum of crash rate magnitudes computed in this study.

\section*{RESULTS}\label{section:results}

\begin{figure}
    \centering
    \includegraphics[width=3.5in]{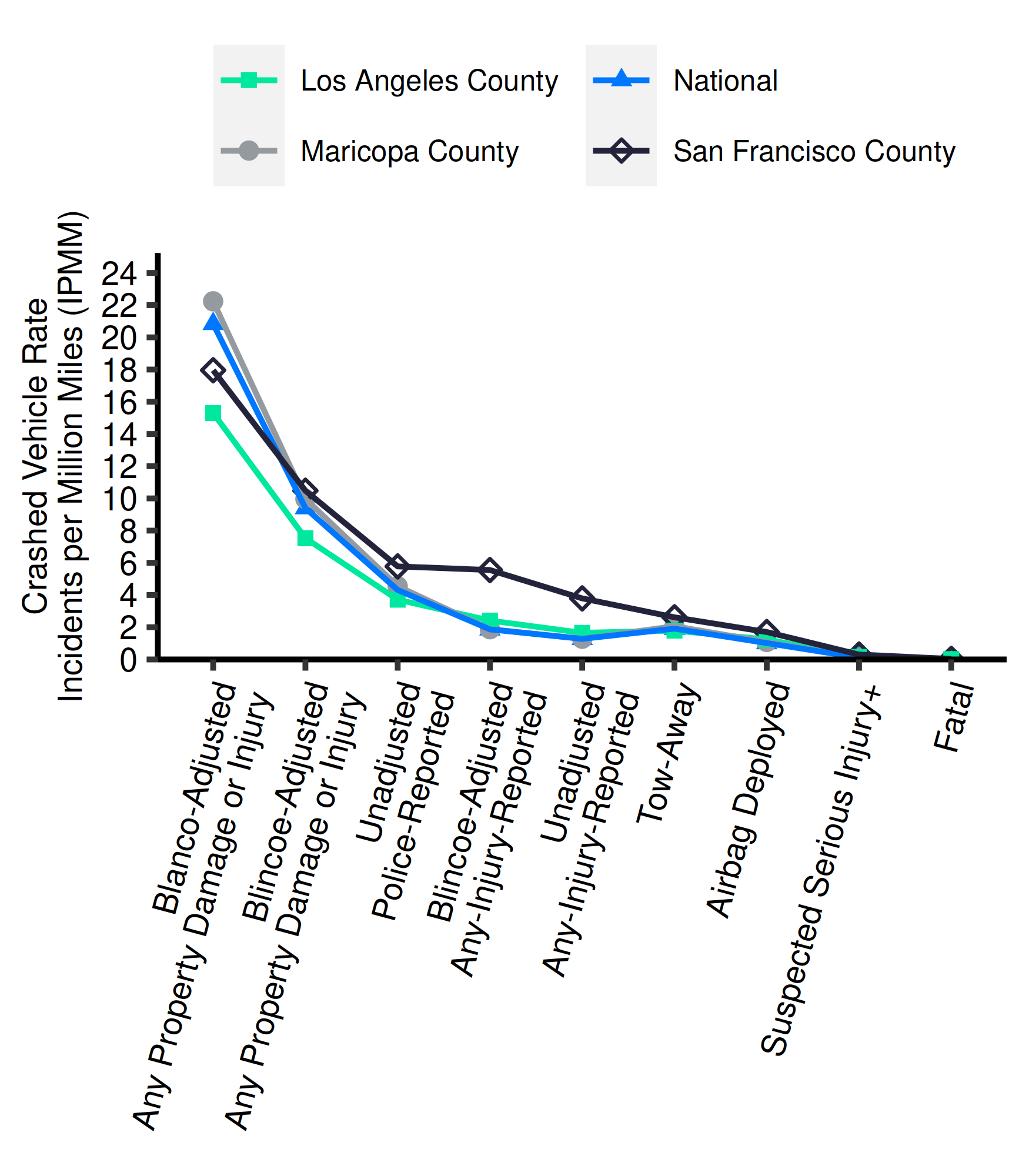}
    \caption{ADS-comparable benchmarks generated for passenger vehicles on surface streets in currently deployed geographic regions. Crashed vehicle rates are presented at multiple severity levels.}
    \label{fig:benchmarking_outcomes_geo}
\end{figure}

Figure~\ref{fig:benchmarking_outcomes_geo} shows this study’s ADS-calibrated benchmarks. Table~\ref{tab:benchmark_results} shows the values of these computed benchmarks with intermediate crash counts, mileages, and crashed vehicle rates, prior to adjusting for road and vehicle type. The relative effects of geographic region, vehicle type, and road type were examined. 

\pagebreak

\newgeometry{margin=1cm} %
\begin{landscape}
\renewcommand{\arraystretch}{1.35}

\begin{table*}[width=8.5in, cols=9, pos=h]
  \caption{A full compilation of the 2022 crash rate benchmarks (presented in crashed vehicle incidents per million/billion Miles, IPMM / IPBM) with intermediate values is presented. Mileage is presented in million miles (Mmi). This includes crash rate tabulations by geographic region and after adjusting for vehicle type and road type.}
  \centering
    \begin{tabular*}{\tblwidth}{|c|r|r|c|c|c|c|c|c|}
      \cline{5-8}
      \multicolumn{4}{c|}{} & \textbf{National} & \textbf{Maricopa County} & \textbf{San Francisco County}	& \textbf{Los Angeles County} \\
      \cline{1-9}
      \multicolumn{2}{|c|}{\multirow{5}{*}{\rotatebox{90}{\parbox{1in}{\centering\textbf{All Police-reported}}}}} & \multicolumn{2}{r|}{\textbf{Mileage}} & 3,196,191 Mmi & 43,095 Mmi & 2,239 Mmi & 71,539 Mmi & \multirow{8}{*}{\rotatebox{270}{\centering\textbf{Intermediate Values}}} \\
      \cline{3-8}
      \multicolumn{2}{|c|}{} & \multirow{2}{*}{\textbf{Crashes}} & \multirow{2}{*}{Unadjusted} & \multirow{2}{*}{5,930,496} & \multirow{2}{*}{85,991} & \multirow{2}{*}{5,809} & \multirow{2}{*}{103,524} & \\
      \multicolumn{2}{|c|}{} & & & & & & & \\
      \cline{3-8}
      \multicolumn{2}{|c|}{} & \multirow{2}{*}{\parbox{1.5in}{\flushright\textbf{All Vehicles, Any Type (IPMM)}}} & \multirow{2}{*}{Unadjusted} & 10,528,849 & 168,619 & 9,482 & 188,504 & \\
      \multicolumn{2}{|c|}{} & & & (3.29 IPMM) & (3.91 IPMM) & (4.24 IPMM) & (2.63 IPMM) & \\
      \cline{1-8}
      \multirow{18}{*}{\rotatebox{90}{\centering\textbf{Crashed Passenger Vehicles}}} & \multirow{3}{*}{\rotatebox{90}{\parbox{0.35in}{\centering\textbf{All Roads}}}} & \multicolumn{2}{r|}{\textbf{Mileage}} & 2,822,666 Mmi & 39,189 Mmi & 1,932 Mmi & 61,846 Mmi & \\
      \cline{3-8}
      & & \multirow{2}{*}{\textbf{Police-Reported (IPMM)}} & \multirow{2}{*}{Unadjusted} & 9,685,661 & 153,819 & 8,492 & 174,070 & \\
      & & & & (3.43 IPMM) & (3.93 IPMM) & (4.39 IPMM) & (2.81 IPMM) & \\
      \cline{2-9}
      & \multirow{15}{*}{\rotatebox{90}{\centering\textbf{Surface Streets}}} & \multicolumn{2}{r|}{\textbf{Mileage}} & 2,140,140 Mmi & 24,865 Mmi & 862 Mmi & 29,952 Mmi & \multirow{15}{*}{\rotatebox{270}{\textbf{ADS Benchmarks}}} \\
      \cline{3-8}
      & & \multirow{4}{*}{\textbf{\parbox{1.5in}{\flushright Any Property Damage or Injury (IPMM)}}} & \multirow{2}{*}{Blincoe-Adjusted} & 19,122,761 & 234,451 & 9,041 & 204,880 & \\
      & & & & (8.94 IPMM) & (9.43 IPMM) & (10.49 IPMM) & (6.84 IPMM) & \\
      \cline{4-8}
      & & & \multirow{2}{*}{Blanco-Adjusted} & 42,432,501	& 522,283 & 15,155 & 413,341 & \\
      & & & & (19.8 IPMM) & (21.0 IPMM) & (17.6 IPMM) & (13.8 IPMM) & \\
      \cline{3-8}
      & & \multirow{2}{*}{\textbf{Police-Reported (IPMM)}} & \multirow{2}{*}{Unadjusted} & 8,768,951 & 107,167 & 5,049 & 101,590 & \\
      & & & & (4.10 IPMM) & (4.31 IPMM) & (5.86 IPMM) & (3.39 IPMM) & \\
      \cline{3-8}
      & & \multirow{4}{*}{\parbox{1.5in}{\flushright\textbf{Any-Injury-Reported (IPMM)}}} & \multirow{2}{*}{Unadjusted} & 2,583,558 & 30,791 & 3,426 & 46,275 & \\
      & & & & (1.21 IPMM) & (1.24 IPMM) & (3.98 IPMM) & (1.54 IPMM) & \\
      \cline{4-8}
      & & & \multirow{2}{*}{Blincoe-Adjusted} & 3,774,625 & 44,933 & 5,015 & 67,622 & \\
      & & & & (1.76 IPMM) & (1.81 IPMM) & (5.82 IPMM) & (2.26 IPMM) & \\
      \cline{3-8}
      & & \multirow{2}{*}{\textbf{Tow-Away (IPMM)}} & \multirow{2}{*}{Unadjusted} & 3,817,957 & 50,742 & 2,351 & 46,318 & \\
      & & & & (1.78 IPMM) & (2.04 IPMM) & (2.73 IPMM) & (1.55 IPMM) & \\
      \cline{3-8}
      & & \multirow{2}{*}{\textbf{Airbag Deployed (IPMM)}} & \multirow{2}{*}{Unadjusted} & 2,070,647 & 32,748 & 1,539 & 34,909 & \\
      & & & & (0.97 IPMM) & (1.32 IPMM) & (1.79 IPMM) & (1.17 IPMM) & \\
      \cline{3-8}
      & & \multirow{2}{*}{\parbox{1.5in}{\flushright\textbf{Suspected Serious Injury+ (IPMM)}}} &
      \multirow{2}{*}{Unadjusted} & 232430 & 2598 & 271 & 4561 & \\
      & & & & (0.11 IPMM) & (0.10 IPMM) & (0.31 IPMM) & (0.15 IPMM) & \\
      \cline{3-8}
      & & \multirow{2}{*}{\textbf{Any Fatality}} & \multirow{2}{*}{Unadjusted} & 38,507 & 602 & 34 & 702 & \\
      & & & & (18.0 IPBM) & (24.2 IPBM) & (39.5 IPBM) & (23.5 IPBM) & \\
      \cline{1-9}
      
    \end{tabular*}
    \label{tab:benchmark_results}
\end{table*}

\end{landscape}
\restoregeometry

\renewcommand{\arraystretch}{1.0}

\pagebreak

\subsection*{ADS-Comparable Benchmarks Across Different Geographic Regions}

ADS comparable rates (crashed passenger vehicles on surface streets) were successfully generated for \textit{any property damage or injury}, \textit{police-reported}, \textit{any-injury-reported}, \textit{tow-away}, and \textit{fatal} crashes. 
The effect of geographic region on the data was notable at multiple severity levels. When examining \textit{Blincoe-adjusted any-injury-reported} vehicle crash rates, all ADS driving regions had higher crash rates than the adjusted national average. San Francisco had the highest crash rate with 5.82 incidents per million miles (IPMM), which was approximately three times higher (330\% higher) than the national average (1.76 IPMM). Maricopa County’s rate (1.80 IPMM) was nearly identical to the national average. Los Angeles county (2.26 IPMM) had a crashed vehicle rate approximately 28\% higher than the national average. 

\subsection*{Vehicle and Road Type Effects}

Passenger vehicles tended to have a higher crash rate than the overall crashed vehicle rate, which
included passenger vehicles, heavy vehicles, and motorcycles. Nationally, on all roadways, the
police-reported IPMM was 4\% higher for passenger vehicles than all crashed vehicles (prior to subsetting
by vehicle type). Also, San Francisco county (4\% higher) and Los Angeles county (7\% higher) each had a higher passenger vehicle crash rate than all vehicle crash rate. Virtually no difference was observed in the crash rate between passenger vehicles and all crashed vehicles in Maricopa county.  

Surface streets tended to have a higher passenger vehicle crash rate when compared to all roadways, which includes highways. This was examined by comparing the police-reported passenger vehicle crash rate before and after adjusting for road type. Nationally, this surface street passenger vehicle crash rate was 19\% higher than the all road crash rate. In all county-level evaluations, the surface street crash rate was higher by a range of 10\% (for Maricopa County) to 33\% (for San Francisco County).

\subsection*{Statistical Power Analysis}

\begin{figure}
    \centering
    \includegraphics[width=3.5in]{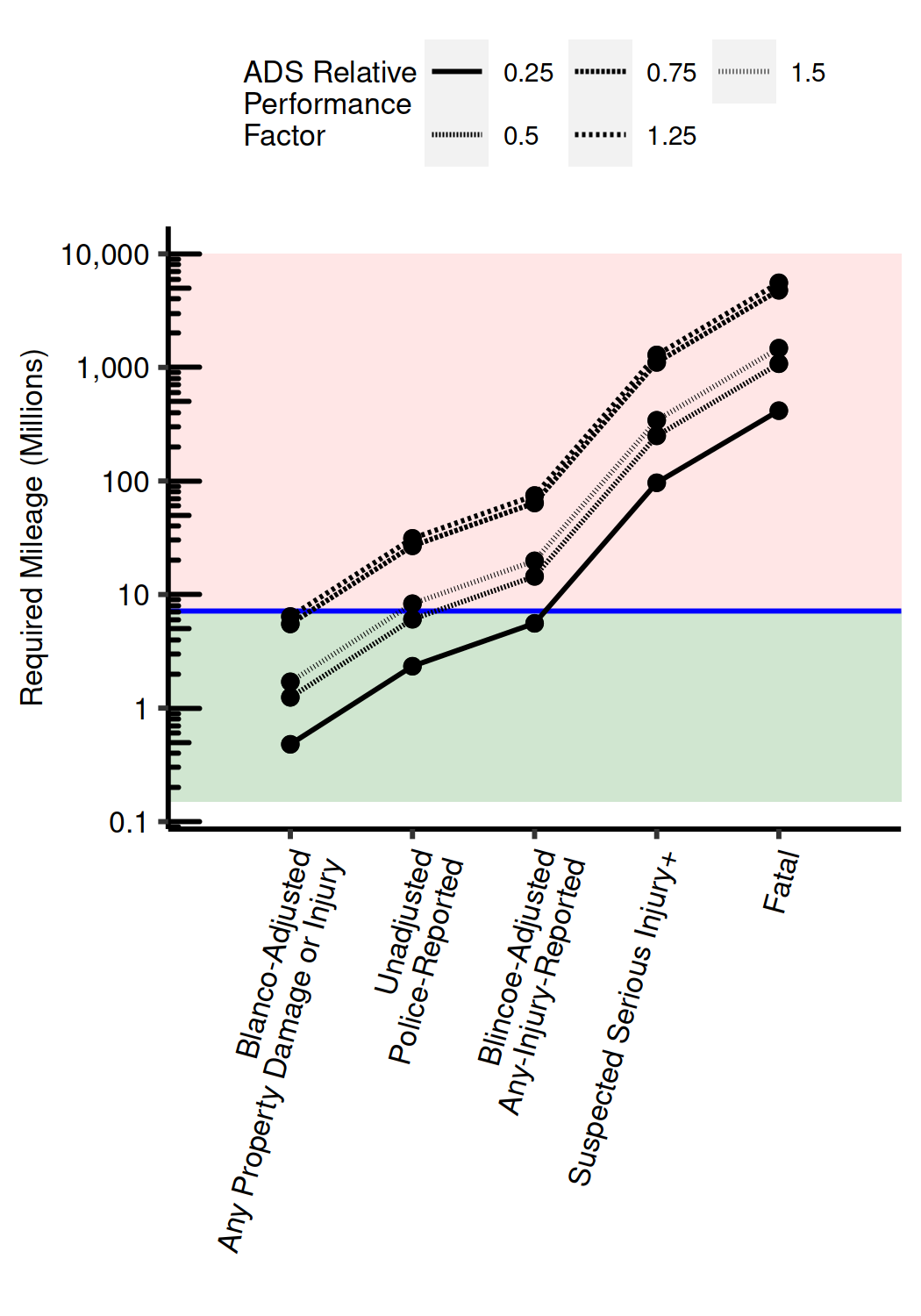}
    \caption{VMT (in millions) required to demonstrate statistically significant crash rate performance for a fictive ADS with 80\% power using a 2-sided test with $\alpha = 0.05$.}
    \label{fig:ads_power}
\end{figure}

Table~\ref{table:power} shows the results of the statistical power analysis for determining the approximate number of miles needed to show a statistically significant difference at an $\alpha = 0.05$ level relative to national benchmarks with 80\% power. Figure~\ref{fig:ads_power} shows the same data as in Table~\ref{table:power} graphically. 

\begin{table*}[width = \textwidth, cols=7]
\caption{VMT (in millions) required to demonstrate statistically significant crash rate performance for a fictive ADS with 80\% power using a 2-sided test with $\alpha = 0.05$.}
\centering
\begin{tabular*}{\tblwidth}{rccccccc}
  \toprule
    & \multicolumn{7}{c}{\parbox{3.0in}{\centering\textbf{Fictive ADS Crashed Vehicle Rate Relative to Human Benchmark}}} \\
    & \textbf{1\%} & \textbf{10\%} & \textbf{25\%} & \textbf{50\%} & \textbf{75\%} & \textbf{125\%} & \textbf{150\%} \\ 
  \midrule
  {\em Blanco-Adjusted Any Property Damage or Injury} & 0.2 & 0.3 & 0.5 & 1.3 & 5.8 & 6.8 & 1.8 \\ 
  {\em Unadjusted Police-Reported} & 1.0 & 1.5 & 2.5 & 6.4 & 28.2 & 32.9 & 8.7 \\ 
  {\em Blincoe-Adjusted Any-Injury-Reported} & 2.4 & 3.5 & 5.7 & 14.8 & 65.6 & 76.3 & 20.3 \\ 
  {\em Suspected Serious Injury+} & 39.3 & 56.3 & 92.8 & 240.4 & 1,065.1 & 1,239.8 & 329.4 \\ 
  {\em Fatal} & 236.9 & 340.0 & 560.0 & 1451.3 & 6429.0 & 7483.3 & 1988.5 \\ 
  \bottomrule
\end{tabular*}
\label{table:power}
\end{table*}

The results show the required mileage needed to determine statistically significant results given some ADS crash rate. Consider, for example, a fictive ADS \textit{suspected serious injury+} crash rate that is observed to be 50\% of the benchmark crash rate. This analysis indicates that approximately 240 million miles driven by the ADS would be required to determine statistical significance. Conversely, if that crash rate was 10\% of the benchmark crash rate, the required mileage would only be approximately 56 million miles. 

Millions of miles are needed to determine statistically significant differences at the \textit{Blanco-adjusted any property damage or injury}, \textit{police-reported}, and \textit{Blincoe-adjusted any-injury-reported} crash severity levels. Fictive ADS crash rates closer to the benchmark rate require more miles to establish statistical significance. For the fatal vehicle crash rates, the required ADS VMT to establish statistical significance  ranges from hundreds of millions to billions of miles of driving. An important conclusion from Table~\ref{table:power} is that both the magnitude of the the benchmark rate and the magnitude of the relative ADS performance difference determines the VMT required for statistical significance. 

\section*{DISCUSSION}\label{section:discussion}

The results provide sets of ADS comparable human crash benchmarks across multiple geographic areas, where SAE level 4 ADS-equipped ride-hailing fleets are currently deployed. Crash rates for several outcome levels were presented. \textit{Any property damage or injury} crash rates were notably the most frequent of the crash rate outcome levels. \textit{Fatal} crash rates were the least frequent. The importance of considering geographic region, vehicle body type, and road type in the computation of benchmarks were demonstrated to be influential factors. Failing to account for each of these features when generating ADS-comparable benchmarks was generally found to cause an overestimation of human driving performance (i.e., would result in a lower-than-actual human crash rate) within the current ADS ODD. 

Researchers face multiple challenges when generating crash rate benchmarks. Each subsequent section details individual challenges identified from a review of the existing literature and in the current benchmark generation exercise. The decisions made in the current study are presented alongside these past studies. Considerations of the limitations of the current study and for future work are also presented. 

\subsection*{Measuring Vehicle Crash Rates}\label{subsection:vehicle_crash_rates}

We identified multiple publications that improperly compared human crash-level rates to ADS vehicle-level rates \citep{banerjee2018hands, blanco2016automated, cummings2024assessing, favaro2017examining, schoettle2015preliminary}. Crash rates derived from ADS data represent the number of times an ADS crashes for some amount of VMT (a vehicle-level or driver-level crash rate). The appropriate benchmark comparison to this ADS rate would also be a vehicle-level crash rate, or the number of times human drivers crashed over some amount of VMT. These past studies used the number of overall crashes (a crash-level rate) rather than the number of involved vehicles (vehicle-level). The total number of crashes is not equivalent to the total number of drivers that crashed, because crashes, more often than not, involve multiple vehicle drivers. For this comparison to vehicle-level rates, using crash counts instead of the number of crashed vehicles leads to undercounting, which specifically underestimates the human crash rate (a better safety performance than in reality) and does not represent the number of human drivers who crashed divided by the mileage they had accumulated. 

We estimate what the effects of this commonly performed error would be using the results from the current study. Using the 2022 national police-reported crashes and VMT in Table~\ref{tab:benchmark_results}, there were 5.9 million police-reported crashes in 2022 over 3.2 trillion VMT. These 5.9 million crashes involved a total of 10.5 million crashed vehicles, which computes to 3.29 police-reported crashed vehicles per million miles (an average of 1.78 drivers in every crash). The result is that the vehicle-level rate is 78\% higher than the crash-level rate using the national crash statistics. So, making this mathematical error when comparing to ADS vehicle-level rates would serve to greatly underestimate ADS performance in a comparison. A more detailed summary of this mathematical error, including additional examples of its occurrence (both correctly and incorrectly) within the literature, can be found in Appendix~\ref{appendix:veh_level_crash_rate}. 

\subsection*{Reporting Thresholds and Underreporting}\label{subsection:reporting_threshold}

Police-reported crash databases have important considerations around reporting thresholds and underreporting. Reporting thresholds determine which crashes are eligible for inclusion. Underreporting indicates which crashes, despite being eligible for inclusion, might still be missing. Failing to make this correction could lead to an incorrect statistical conclusion when comparing to ADS data if the reporting threshold and/or underreporting in the ADS data is different than in the human data. This particular work focuses on developing benchmarks that can be compared to ADS crash rates. ADS operators are required to report a crash involving any ``physical impact'' that ``results or allegedly results in any property damage, injury, or fatality'' as a part of the NHTSA SGO reporting \citep{nhtsa2023b}. For human police-reported data, crashes are reported to and by police if they meet a jurisdiction-specific reporting threshold. These thresholds can vary widely. For example, California and Arizona have a \$1,000 and \$300 reporting threshold, respectively, while states like Pennsylvania require a vehicle to have been towed away from the scene \citep{blincoe2023economic}. Police-reportable crashes (i.e., crashes where reporting thresholds are met) are also often simply not reported to the police in human driver only crashes. There are many reasons for this underreporting to police \citep{mdavisandco}. Even if a crash is reported, a judgement call is still being made by a police officer as to whether the reporting threshold was met. 

There is an additional source of crash underreporting for police-reported crashes in California. The SWITRS database (California’s state crash database), by design, is intended to capture all reported injuries and fatalities. There is no requirement to submit PDO crashes, and, accordingly, ``some agencies report only partial numbers of their PDO crashes, or none at all'' \citep{chp2021}. There is no explicit indication in the available literature from the state on whether this underreporting limitation applies to Los Angeles and San Francisco counties. An initial investigation, presented in Appendix~\ref{appendix:switrs_underreporting}, suggested that there were substantially fewer PDO crashes than injury crashes in these California counties with respect to what is observed in Maricopa County and the national average. This result indicates that a potentially large number of reported PDO crashes were not incorporated in the SWITRS database, but more investigation is needed. 

This study employed two different strategies for accounting for these aforementioned reporting threshold and underreporting challenges. The first strategy was to consider only crash severity levels where reporting thresholds can be matched and underreporting effects were minimal. The second strategy was to estimate the data missing due to reporting threshold differences and underreporting. 

\subsubsection*{Matched Reporting Thresholds}\label{subsection:reporting_threshold}

Multiple benchmarks were computed by taking subsets of the police report data (e.g., \textit{suspected serious injury+}, \textit{any-injury-reported}, \textit{airbag deployment}). These subsets of police-reported outcomes are advantageous for two reasons. First, reporting thresholds can be better matched between the benchmark and ADS data compared with extrapolating the police-reported data using underreporting adjustments. Second, underreporting concerns are reduced at higher severity levels due to involved vehicles sustaining substantial property damage and/or involved persons requiring emergency medical attention \citep{blincoe2015economic, blincoe2023economic}.

Restricting analyses to a matched ADS and benchmark reporting threshold has, to the knowledge of the authors, only been employed in one prior ADS benchmarking analysis using police report data. \citet{teoh2017rage} reviewed individual case narratives from California (CA) Department of Motor Vehicle (DMV) OL 316 reported crashes, which are reported by the ADS operators \citep{cadmv2023}, to determine whether each event would have met the California police reporting criteria. These identified crashes were then used to generate a benchmark ``police-reported'' crash rate. Although this reporting threshold was considered, no correction due to underreporting effects was employed in \citet{teoh2017rage}. 

It is notable that higher severity injury outcome crashes have been documented to still have some degree of underreporting. \citet{blincoe2023economic} estimated that 27\% and 6\% of moderate (MAIS2) and seriously injured (MAIS3) persons, respectively, are not reported to police. In another study, the San Francisco Department of Public Health (SFDPH) collaborated with the San Francisco Municipal Transportation Agency (SFMTA) on a ``Vision Zero SF'' project examining high injury locations within San Francisco \citep{sfhealth2017}. Their work developed a comprehensive Transportation-related Injury Surveillance System (TISS) to monitor reported injuries in both police-reported data and through hospital records. There was specific emphasis on ``severe'' injuries, which were defined by presentation of several predefined injuries, any visit to a hospital trauma center, or an injury severity score (ISS) of 15 or greater. When examining hospital records against police-reported data, their analysis found 39.2\% of vehicle occupants, 27.7\% of pedestrians, and 33.1\% of cyclist riders with severe injuries caused by motor vehicle crashes did not have an associated police report by the San Francisco Police Department.  

\subsubsection*{Crash Rate Adjustments}\label{subsection:crash_rate_adjusments}

The second strategy to consider is adjusting the crash data to account for mismatched reporting thresholds and underreporting effects. This study used underreporting estimates from \citet{blanco2016automated} and \citet{blincoe2023economic} to generate three adjusted crash rates: \textit{Blanco-adjusted any property damage or injury}, \textit{Blincoe-adjusted any property damage or injury}, and \textit{Blincoe-adjusted any-injury-reported} crash rate. 

These techniques are not novel to the ADS benchmark literature. Both \citet{schoettle2015preliminary} and \citet{blanco2016automated} generated crash rates similar to the \textit{any property damage or injury} level used in the current study. \citet{schoettle2015preliminary} adjusted police-reported crash rates according to NHTSA's underreporting estimates for PDO and non-fatal crashes that were presented in \citet{blincoe2015economic}. \citet{blanco2016automated} generated adjusted police-reported crashes using both the \citet{blincoe2015economic} non-fatal crash rates and their own analysis of SHRP-2 data.  

There were stark differences in the computed \textit{any property damage or injury} adjusted crash rates, where the Blanco-adjusted rate was consistently close to or over two times the Blincoe-adjusted crash rate. This is likely a reflection of multiple factors, but, generally speaking, both adjustment techniques have important limitations to consider that affect these correction factors. 

The \citet{blincoe2015economic} estimates were generated using a combination of telephone surveyed data, insurance records, and other sources. In the telephone survey, respondents were asked to self-report crashes that resulted in any type of damage with no guidance on a minimum threshold. \citet{blincoe2023economic} was also designed to be nationally-representative and relied on data and analysis from over a decade ago. Underreporting effects change over time and have geographic dependence \citep{blincoe2015economic}. 

\citet{blanco2016automated} did not assess vehicle damage directly. Rather, the investigators relied upon video and sensor analysis to determine ``police-reportable'' status (based on a damage threshold of \$1500), which introduces some potential bias and is notably not higher than ``any property damage.'' \citet{blanco2016automated} investigators also relied on self-reporting by participants to determine whether or not a police report was filed, which introduced potential undercounting of reported crashes. SHRP-2 data was also not collected from any of the specific geographic regions being investigated in this study, which raises questions about how representative this estimate may be. Overall, the biases in \citet{blanco2016automated} work in both the conservative and anti-conservative directions and have unknown magnitude.

\subsubsection*{Selecting Benchmarks for ADS Evaluation}\label{subsection:benchmark_selection}

When these or similar benchmark rates are used, it is essential that researchers properly contextualize the results with respect to their unique limitations. Our recommendations on using the current study's benchmarks are presented below. 

The \textit{any property damage or injury} benchmark has arguably the highest uncertainty of all investigated benchmarks. Using these adjusted crash rates, or similarly adjusted rates, should be done with caution. Although the underreporting and reporting threshold issues are both being addressed, the \citet{blincoe2015economic, blincoe2023economic} and \citet{blanco2016automated} adjustments are inconsistent with each other and not necessarily representative of current case years and geographic areas being investigated. Another limitation is that the data used to estimate underreporting in these previous studies does not have a clear lower reporting threshold. This lack of lower reporting threshold contributes to the large differences observed in the adjusted police-reported data and NDS crash rates discussed previously. Future research should focus on developing benchmarks with a clear lower reporting threshold that can more precisely represent this \textit{any property damage or injury} outcome group.  

\textit{Police-reported} and \textit{unadjusted any-injury-reported} crash rates theoretically have a fixed reporting threshold (for a given jurisdiction) but are both known to suffer from underreporting. This makes these benchmark rates underestimate the true \textit{any-injury-reported} crash rates and rates of eligible \textit{police-reported} crashes (i.e., police reportable). Using these estimates in an ADS comparison would generally reflect conservatism in the analysis, which, when stated as such, makes it appropriate to rely upon. 

The \textit{Blincoe-adjusted any-injury-reported} crash rates have improved reliability over the \textit{unadjusted any-injury-reported} estimate. Both estimates apply an identical reporting thresholds. The underreporting magnitude in injury crashes is lower than that in \textit{any property damage or injury} crashes, thus the underreporting adjustment will have a smaller effect on the results. Studies should consider using both the \textit{Blincoe-adjusted any-injury-reported} and \textit{unadjusted any-injury-reported} benchmarks in comparison to ADS crash rates to bound the safety impact estimates for injury outcomes.

We recommend that researchers preferentially consider using benchmarks representing higher severity outcomes. \textit{Any airbag deployment}, \textit{suspected serious injury+}, and \textit{fatal} crash rates were all produced in this study and have greatly reduced underreporting concerns. Reporting thresholds can be generally be matched for each of these levels.

Although underreporting concerns are lower for \textit{tow-away} crash rates, this benchmark rate is  not recommended due to potential reporting threshold challenges when comparing to an ADS driving crash record. Human driven vehicles are generally towed due to disabling vehicle damage, where a person is physically unable to operate the vehicle. ADS-equipped vehicles may sustain only minor damage but still require a tow due to a variety of reason, such as logistics or compromised sensors.  

\subsection*{Accounting for Influential Factors}\label{subsection:driver_matching}

Driving context influences crash and injury rates. This study takes multiple steps to reduce bias resulting from vehicle body type, geographic area, and road type, which are all well known factors influencing crash risk \citep{kmet2006urban, choudhary2018impacts, kweon2005safety}. The current ADS fleets are implemented on passenger vehicle platforms, restricted to specific geographic regions, and are operating on surface street roadways. These three influential factors were selected, alone, because they represent the common denominator across the available mileage exposure and crash data set relied upon. Specifically, the state mileage data had only two dimensions available: road type and geographic location (at the county-level). The national FHWA data compiled for each state was required to add in the vehicle-level correction. All three of these factors were found to influence crash rates in the current study, which further highlight their importance in incorporating into the benchmark rates estimates.

To our knowledge, no prior study had considered all three features when generating benchmarks from police-reported data. The majority of prior ADS benchmarking studies using police report data considered only national crash data without any sort of geographic correction \citep{banerjee2018hands, dixit2016autonomous, favaro2017examining, cummings2024assessing}. None of the existing police-reported benchmark studies considered Maricopa, Los Angeles, and San Francisco counties, where a large proportion of ADS driving miles in the United States are currently taking place. A road type adjustment had been implemented by both \citet{teoh2017rage} and \citet{cummings2024assessing}. 

A recent analysis by \citet{flannagan2023establishing} used a 5.6 million mile NDS dataset (mostly of ride-hailing drivers) taken from San Francisco roadways to examine crash rates. This study appears to be the first NDS human benchmark analysis taken directly from the San Francisco geographic area. Their study used a combination of naturalistic driving records and insurance information to identify crashes. The reporting threshold was generally unclear, i.e., how the combination of kinematics triggers and insurance claims were leveraged to determine what should and should not be counted as an event. \citet{flannagan2023establishing} estimated that the NDS fleet experienced 64.9 crashes per million miles, which is multiple times higher than the \textit{any property damage or injury} crash rate estimated in the present study for San Francisco (10.5 to 17.6 IPMM). Without establishing a set reporting threshold, it is not clear if  this differences are due to underestimation using the reporting adjustments in the current study, a lower inclusion criteria relied upon in the \citet{flannagan2023establishing} study, or some combination of both.    

Even after accounting for these influential features, there still remains confounders. The effect a potential confounder has on a result is dependent on (a) how different the exposure is between the benchmark and ADS data and (b) how influential that feature is (effect size) on crash risk. The driving exposure of current ADS deployments are not inherently equivalent to the exposure of the benchmark crash population. For example, ride-hailing fleets may operate more at night, in areas of denser traffic, or areas with more vulnerable road users than the general driving population. Relevant features, such as time-of-day, weather, day of the week, vulnerable road user presence, and traffic density, can all influence crash rates \citep{martin2002relationship, regev2018crash, qiu2008effects}. Future work should consider additional data and analytical steps to help account for known, potentially influential features.

\subsection*{Other Relevant Data Considerations}

\subsubsection*{Risk-based Measures}

Risk-based measures of safety should be considered as an alternative to outcome measures. Prior works, including the current study, have generally demonstrated the utility of ``outcome'' data in retrospective safety assessment. Outcomes, such as injuries, vehicle damage, or property losses, are highly dependent on individual or vehicle characteristics (e.g., person age, sex, seat belt use) and, to a certain degree, chance. The safety community has historically considered injury risk-based assessments as an alternative approach to measuring safety impact \citep{campolettano2023representative, kullgren2008dose, Kusano_ITS, KusanoCAT, Scanlon_TIP_IADAS, scanlon2022collision}. Using this approach, the dynamics of the crash are used to assess injury risk \citep{McMurry2021, schubert2023passenger,lubbe2022safe}. The benefits of this approach is (a) it can provide earlier indication of safety performance on injury-relevant crashes and (b) crash prevention and mitigation can be more explicitly measured by controlling for other influential factors.

\subsubsection*{Exclusion Bias}

Studies should also be careful not to unintentionally introduce exclusion bias. This bias can arise when a type of outcome is excluded from the data, as occurs in \citet{cummings2024assessing}, where fatal crashes were excluded. Because fatal crashes were removed from the event-numerator, all fatal crashes were, in effect, re-classified as trips where no crash occurred. These biases may not be a large influence on the results due to the relative rarity of fatal crashes, but they remain important conceptual challenges. 

\subsubsection*{Additional Analytical Lenses to Consider}

The current study focused on the general crash population for generating benchmarks. Comparing ADS to the general crash population enables an estimate on the net effect of the introduction of the technology on the current population of drivers. Data permitting, researchers should consider additional benchmarks to further explore safety impact. \citet{flannagan2023establishing} provides an example of more narrowly focusing on ride-hailing human drivers, which may have more similar driving characteristics to ADS ride-hailing vehicles than the general population. 

Subsets of the overall driving population representing a higher standard of driving should also be considered. For example, a ``consistently performing, always-attentive driver'' has been demonstrated as a useful high performance bar in ADS assessment of response to imminent high potential severity conflicts \citep{scanlon2022collision}. \citet{goodall2021a} presented a technique for adjusting crash benchmarks for an ``error and impairment-free model driver.'' This idealized ``model driving'' crash rate created a benchmark of human driving superior to the overall driving population that exists only during periods of driving (i.e., ``sober, rested, attentive, cautious''). There are also other opportunities for better performance crash rates, such as comparing to newer vehicles equipped with the latest safety features. Idealized crash rate benchmarks provide additional evaluation criteria that may more closely match societal expectations for driving competency, but do not accurately estimate the effect of a technology on the status quo. Additional research is required to establish additional benchmarks beyond overall status quo benchmarks. 

It is important to recognize that aggregate performance benchmarks, like the ones generated in the current study, include exposure to a variety of driving conditions and crash types. As mileage continues to grow, researchers’ ability to evaluate more granular components, such as narrowly defined crash types or adverse weather conditions, will be enabled. For example, \citet{teoh2017rage} performed an analysis on the Waymo vehicle's (formerly Google Self Driving Car) crash performance by crash type during testing operations (i.e, with a person behind the wheel), which identified differences in their distribution when compared to the human crash population. Similar future analyses, applied to RO driving, will provide important findings on the direction and magnitude of any safety impact achieved by operational services. 

\subsubsection*{Handling Data Missingness}

Crash data is also subject to a certain degree of missingness in the data that is compiled for a given crash. Factors like missing records or a vehicle driving away from the scene of a crash result in certain crash variables not being coded, such as the type of vehicle, injury status, or damage extent. Researchers should consider various strategies for handling missingness, such as relying on imputed data, when handling police-reported data, and should be aware of how that missingness may introduce bias in the results \citep{McMurry2021}. Given that vehicle type was occasionally not specified with enough detail, this study relied on a form of imputation for estimating the passenger vehicle crash counts. 

\subsubsection*{Year-to-Year Effects}

Another consideration is that the current study only considered crash and mileage data from 2022. This was the most recent data present across all data sources. Crash rates do vary from year-to-year \citep{stewart2023overview}. An analysis of prior years can be seen in Appendix \ref{appendix:year_effects_crash_rates}, where the year-to-year dependence is demonstrated with this study's benchmarks. It was observed that the year-to-year crash rate trends are dependent on both severity level and geographic region. Consistent with previous work, there were clear trends observed around 2020, which was during the COVID-19 pandemic \citep{grembek2022highway, meyer2020covid}. The crash rates remained relatively stable between 2021 and 2022. The year-to-year trends will need to be revisited as new data is released. 

Using this study's results from 2022 data to compare to other ADS driving years should be done with caution. Ideally, researchers would match ADS crash rates for some time period to some benchmark rates from the same period, but there is a lag in publishing police reported crash data. In general, national and state police-report databases are published for each complete calendar year, and are typically published several months to years after the calendar year is over. In comparison, NHTSA SGO data for ADS crashes are reported monthly. The results presented from this study should be revisited using future years of data as they become available. 

\subsection*{Safety Impact Considerations for High Severity Crash Outcomes}

The statistical power analysis performed in this paper is not unlike the analysis shown in Figure 3 of \citet{kalra2016driving}. A detailed accounting of differences between this study and this prior work can be found in Appendix~\ref{appendix:kalra_paddock}. To summarize, this previous study used a fatality rate (fatally injured persons per VMT) rather than a vehicle-level crash rate, used an earlier year of crash data, considered a static effect size, and considered a generic, national driving exposure. Generally speaking, \citet{kalra2016driving} find a similar order of magnitude of necessary ADS miles to determine statistical significance than presented in the current study. 

The statistical power analysis performed in this study leveraged national crash and VMT estimates to determine approximate ADS VMTs where statistically significant differences are likely to be observed given various effect sizes. Two contributors of statistical randomness were considered in this analysis: (1) the inherent randomness around the mean in the number of observed ADS events and (2) uncertainty in the estimated human-driven non-fatal crash totals due to the sampling structure in CRSS. There are other forms of uncertainty that are unaccounted for, but that should be considered when interpreting these results. First, there is uncertainty in the transformed  estimates of \textit{Blanco-adjusted any property damage or injury} and \textit{Blincoe-adjusted any-injury-reported} due to the adjustments that were performed. For this current study, we considered these adjustments as point estimates only. Second, the national VMT estimates are, themselves, the product of sampling and estimation by the various local, state, and federal agencies. Accounting for these additional uncertainties would lead to a higher estimated VMT required for detecting statistical significance.

\section*{CONCLUSIONS}\label{section:conclusion}

This paper presents human benchmarks for multiple geographic regions and outcome levels that are comparable to available ADS crash data sources to help advance the retrospective assessment of ADS technology. ADS fleets are increasing miles driven, and researchers are beginning to have an opportunity for more robust statistical evaluation of safety impact. With that opportunity, the safety research community needs to be vigilant about implementation of benchmarks that are comparable to ADS crash data. A review of the existing literature highlights multiple opportunities for error and bias. These common challenges are identified, their effect is quantified, and a methodological solution is presented, which serves to inform the automotive safety community on careful, conservative practices that can be used for crash rate benchmark generation. This study demonstrated that police report data can used to generate benchmarks for various outcomes, but researchers should be cautious when comparing any \textit{property damage or injury}, \textit{police-reported}, and \textit{any-injury-reported} rates against observed ADS performance, which each have underreporting limitations that must be corrected in analysis. Features like geographic region, vehicle type, and road type all play a role in influencing crash rates. The presented benchmarks are only one iteration. Researchers, regulators, industry, and experts should consider these results, and have an opportunity to continue to elevate the state-of-the-art of benchmarking. 

\section*{Disclosure Statement}\label{section:disclosure}

All authors are employed by Waymo LLC.

\bibliographystyle{cse_author_name}

\bibliography{refs}

\appendix

\section{APPENDIX}

\subsection{Historical ADS Benchmarking Efforts}

\subsubsection{Review of human crash benchmark data sources}
\label{appendix:crash_data_sources}

Benchmark creation efforts - for both ADS and ADAS - have largely considered three primary data sources: Insurance, naturalistic driving study, and police-reported. When examining the performance of an ADS, evaluators generally want to understand, for the benchmark, how often does that population of drivers crash given some amount of driving exposure? Generating this vehicle-level crash rate estimate requires a count of total crashed vehicles (or crash involved drivers) and an estimation of the total VMT. It is important to consider: (a) which population of drivers are ADS comparable (e.g., geographic region, vehicle type, road types)? and (b) what crash outcomes is of interest? Data source selection and methodological choices around these questions should be driven by the desired research question. 

Insurance data has served as one credible source for retrospectively assessing performance \citep{cicchino2017effectiveness, isaksson2015evaluation, isaksson2018evaluation}. Data tables derived from insurance claims data have several key advantages for measuring safety impact. First, there are established insurance reporting practices, which help to limit surveillance bias due to reporting thresholds \citep{dilillo2023comparative}. Second, claims data tends to capture a higher proportion of safety-relevant crash events (better event recall) than police-reported data \citep{isaksson2018evaluation,blincoe2023economic}. One challenge with relying on insurance data is driving mileage is generally not directly measured, and other data sources are required to estimate driving volume and mix (e.g., exact locations, time-of-day, weather). Insurance data usage in ADAS retrospective safety impact has largely been restricted to analysis only after substantial mileage has been collected. 

The Swiss Re Group (Swiss Re) recently led a study examining Waymo’s third party liability claims frequency over the company’s first 3.8 million RO miles \citep{dilillo2023comparative}. Third party liability claim rates are distinct from overall crash rates. A third party liability claim occurs only when there is a request for compensation due to property damage or injury by another involved party. Because of this, crashes without attributed liability are excluded from both the benchmark and ADS rates. SwissRe was able to leverage a dataset of over 600,000 claims received over an estimated 125 billion vehicle miles traveled in the areas where Waymo is deployed. Despite a small number of ADS miles compared to the insurance benchmark, this analysis was able to detect a statistically significant reduction in third party property damage and bodily injury claims rate relative to a mileage and zip-code matched human driver benchmark. The detectable difference on a relatively small amount of ADS data (3.8 million miles) was possible because (a) the confidence intervals on the Swiss Re benchmark and ADS data were sufficiently small and (b) the magnitude (effect size) of the ADS-human driver difference was large.

Naturalistic driving study (NDS) data provides another unique opportunity for retrospective safety evaluation, because researchers are able to directly attribute the driving miles to specific geographic areas and can capture events that might not be reported to insurance or police \citep{goodall2021a, blanco2016automated,flannagan2023establishing}. Having a direct measure of the driving provides unique analytical opportunities when looking at how the composition of driving exposure affects crash rate. NDS data has a greater ability to detect and report lower severity events compared to other data sources. On the other hand, NDS data currently published generally covers few geographic regions and only for a limited amount of mileage. \citep{flannagan2023establishing} generated benchmarks for low-severity crashes using a small-scale (approximately 5 ½ million miles) NDS in San Francisco. This study used a combination of insurance and telematics data and recorded NDS data to estimate a property damage crash rate. \citep{blanco2016automated, goodall2021a,goodall2021b} both previously relied upon data collected from the Second Strategic Highway Research Program Naturalistic Driving Study (SHRP‑2) to generate benchmarks. SHRP-2 comprises approximately 34 million driving miles across multiple geographic regions in the U.S. Despite their limited size, the UMTRI study and SHRP-2 dataset enable the analysis of mostly low severity contacts, given how frequently these contacts occur in dense-urban driving environments. However, relying on this data volume alone is limited for measuring safety impact, because lower mileage increases the crash rate estimate confidence intervals. In practice, lower quoted mileage (i.e., on the order of millions) generally available in NDS data limits the ability to make statistical claims around higher severity crash outcomes, such as those resulting in fatality or in moderate, serious, or worse Abbreviated Injury Scale (AIS) injuries \citep{aaam2016}. 

The third commonly relied upon data, and the focus of the current study, is publicly-accessible, police-reported crash databases. The National Highway Traffic Safety Administration (NHTSA) is the primary compiler of national crashes in the U.S., and many states additionally compile their own sets of crashes, which are inherently dependent on the reporting of said crashes to local jurisdictions and then to state-level compilers. Because police crash report data is publicly-accessible, a wider group of researchers are able to leverage data, replicate results, and build upon the current state-of-the-art. These databases take a number of sampling strategies, ranging from stratified sampling to a complete census. For appreciation of scale, the collected annual datasets are generally representative of billions to trillions of miles driven. Where insurance data is generally focused and compiled for a cost-based use case, police-reported data is compiled for quantifying the measurement of injuries, which makes its data features (e.g., KABCO, AIS, crash reconstructions) well suited for safety impact evaluation. Given its accessibility (publicly available) and scale, police-reported data has long served as a primary source of high severity crash rates for measuring safety impact \citep{strandroth2012effects, glassbrenner2009lives}. Like insurance claims data, police-reported crash data does not directly contain associated driving mileage, so other data sources are required to help estimate driving exposure.

\subsubsection{Previous benchmarking efforts using police-reported data}
\label{appendix:prev_benchmark_studies}

Several studies have generated benchmarks for ADS evaluation using police-reported crash and publicly-accessible mileage data. Many of these studies leveraged testing operations data, where a human was seated in the driver seat and had the ability to take control prior to any crash event. Accordingly, observed testing operations crash rates - and computed safety impact - are not necessarily reflective of RO operations. \citet{dixit2016autonomous} computed a California statewide police-reported crash rate using non-fatal injury state crash data (no mileage data source was provided) and compared it to all California (CA) Department of Motor Vehicle (DMV) OL 316 reported crashes, which are reported by the ADS operators \citep{cadmv2023, young2021critical}. \citet{favaro2017examining} and \citet{banerjee2018hands} used national crash data and driving mileage to estimate an overall national crash rate, which they then compared to CA DMV OL 316 reported ADS testing operations crash rates. \citet{schoettle2015preliminary} used national crash databases to derive a benchmark, and adjusted the crash rates using underreporting estimates by a NHTSA study \citep{blincoe2015economic}. Their adjusted benchmark rates were then compared to CA DMV OL 316 reported ADS testing operations crash rates. \citet{blanco2016automated} leveraged state and national police-reported data to generate benchmarks, and used SHRP-2 data and the \citet{blincoe2015economic} study to correct for underreporting in police-reported databases. \citet{blanco2016automated} used these adjusted police-reported crash rates to then compare the benchmark to CA DMV OL 316 reported crashes involving an ADS. \cite{teoh2017rage} used both state and national datasets to generate benchmark rates of police-reported crashes and compared these rates to ``police-reportable'' Waymo driving crash events in testing operations. \citet{cummings2024assessing} used national crash and mileage data to estimate a non-interstate ADS benchmark and then compared the benchmark rates to ADS RO and testing operations crash rates with any injury or property damage, which were gathered from NHTSA Standing General Order (SGO) 2021-01 reporting \citep{nhtsa2023b}. \citet{chen2024} leveraged human crash reporting from Transportation Network Companies (TNCs) with ridehailing operations (for example, Uber and Lyft) to the California Public Utilities Commision (CPUC) to benchmark ADS RO performance from SGO reported crashes. 

A key contribution of this current study is the methodological handling and presentation of these key challenges that researchers face when leveraging this publicly accessible data, which can ultimately unreasonably bias results if not properly handled. Relevant ADS benchmarking literature with features of the works can be found in Table~\ref{tab:prior_studies}. \citet{young2021critical} previously performed epidemiological analysis on six of these prior studies using police-reported data: \citep{schoettle2015preliminary, blanco2016automated, dixit2016autonomous, teoh2017rage, favaro2017examining, banerjee2018hands}. \citet{young2021critical}  extensive review of the prior research noted multiple biases that can be introduced through data and analytical choices that included the aforementioned challenges.

\begin{table*}[width=6.5in, cols=8, pos=h]
  \caption{A compilation of prior studies leveraging human benchmark and ADS data are presented. Studies looking at both ADS testing operations (TO; with a human behind the wheel) and rider-only (RO; without a human behind the wheel or remotely) are included. A study was included only if human and ADS crash rates were compared. The type of ADS data being evaluated, the correct calculation of “vehicle-level” rates (discussed later in the paper), and the presence of relevant data alignment procedures are documented.}
  \centering
  \begin{tabular*}{\tblwidth}{|P{1.5in}|P{0.6in}|P{0.75in}|P{0.5in}|P{0.5in}|P{0.4in}|P{0.4in}|P{0.4in}|}
    \cline{1-8}
    \multirow{2}{1.5in}{\textbf{Study}} & \multirow{2}{0.6in}{\centering\textbf{ADS Data Deployment Stage}} & \multirow{2}{0.75in}{\centering\textbf{Vehicle-Level Benchmark Crash Rates$^1$}} & \multicolumn{5}{|c|}{\textbf{Any ADS \& Benchmark Data Alignment Procedure}} \\
    \cline{4-8}
   & & & \textbf{Geo-Corrected} & \textbf{Reporting Threshold} & \textbf{Vehicle Type} & \textbf{Road Type} & \textbf{Other$^5$} \\
    \cline{1-8}
    \multicolumn{8}{|l|}{\textit{Insurance Data}} \\
    \cline{1-8}
    \citet{dilillo2023comparative} & TO \& RO & \checkmark & \checkmark & \checkmark & \checkmark & $\times$ & $\times$ \\
    \cline{1-8}
    \multicolumn{8}{|l|}{\textit{Naturalistic Driving Study Data}} \\
    \cline{1-8}
    \citet{blanco2016automated} & TO & \checkmark & \checkmark & \checkmark & \checkmark & $\times$ & $\times$ \\
    \cline{1-8}
    \citet{goodall2021a} & TO & \checkmark & \checkmark & \checkmark & \checkmark & $\times$ & $\times$ \\
    \cline{1-8}
    \citet{flannagan2023establishing} & RO & \checkmark & \checkmark & \checkmark & \checkmark & \checkmark & $\times$ \\
    \cline{1-8}
    \citet{chen2024} & RO & \checkmark & \checkmark & $\times^2$ & \checkmark & \checkmark & $\times$ \\
    \cline{1-8}
    \multicolumn{8}{|l|}{\textit{Police-Reported + Public Mileage Data}} \\
    \cline{1-8}
    \citet{banerjee2018hands} & TO & $\times$ & $\times$ & $\times$ & $\times$ & $\times$ & $\times$ \\
    \cline{1-8}
    \citet{dixit2016autonomous} & TO & $\times$ & $\times$ & $\times$ & $\times$ & $\times$ & $\times$ \\
    \cline{1-8}
    \citet{favaro2017examining}$^3$ & TO & $\times$ & $\times$ & $\times$ & $\times$ & $\times$ & $\times$ \\
    \cline{1-8}
    \citet{schoettle2015preliminary} & TO & $\times$ & $\times$ & \checkmark & $\times$ & $\times$ & $\times$ \\
    \cline{1-8}
    \citet{blanco2016automated} & TO & $\times$ & \checkmark & \checkmark & $\times$ & $\times$ & $\times$ \\
    \cline{1-8}
    \citet{teoh2017rage} & TO & \checkmark & \checkmark & \checkmark & $\times^4$ & $\times^4$ & $\times$ \\
    \cline{1-8}
    \citet{cummings2024assessing} & TO \& RO & $\times$ & $\times$ & $\times$ & $\times$ & \checkmark & $\times$ \\
    \cline{1-8}
    \begin{minipage}[c]{0.2\textwidth}\centering Scanlon et al. (2024) \&\newline Kusano et al. (2024)\end{minipage} & RO & \checkmark & \checkmark & \checkmark & \checkmark & \checkmark & $\times$ \\
    \cline{1-8}
  \end{tabular*}
  \raggedright
  \footnotesize{$^1$ Unlike the data alignment procedures, failure to generate ``vehicle-level'' benchmark crash rates invalidates the comparison to ADS crash rates (which are at the vehicle-level). This common mathematical error is discussed further in the paper.}
  
  \footnotesize{$^2$ \citet{chen2024} present a generic "crash" rate comparisons without a severity inclusion threshold and correctly indicate that there is likely relative underreporting of human crash rates when compared to ADS rates. Some analysis of crashes with an injury reported was performed, which would provide some reporting threshold control. A comparison of \textit{Any-injury-reported} rates was not presented though.} 
  
  \footnotesize{$^3$ It is notable that, of the studies presented, \citet{favaro2017examining} did not draw any safety impact conclusions regarding ADS efficacies. Because a human-relative rate was presented in relation to ADS TO crash rates, this study was included.}
  
  \footnotesize{$^4$ \citet{teoh2017rage} do some vehicle type and road type corrections, but do not apply equivalent procedures to both the crash and mileage data. For mileage, they included all vehicle types and only city maintained roads, which excluded mileage on various state highway maintained roads like freeways and interstate 101 passing through city limits. For crashes, they included only passenger vehicles and did not exclude based on road type.}
  
  \footnotesize{$^5$ Additional features, including weather, traffic density, and time-of-day, are known to affect crash rates, but were not accounted for in any of these prior studies, mostly because of lack of mileage data that include these dimensions.}
  \label{tab:prior_studies}
\end{table*}

\subsection{Identifying a Comparable Population of Drivers}
\label{appendix:comp_pop_of_drivers}

\subsubsection{Surface streets}
Ideally, surface streets would be identified using identical encodings across all crash and mileage datasets. Each mileage and crash table, however, have their own unique set of parameters from which road type can be identified. One convenient lens for identifying non-highways is FHWA’s highway function classification coding \citep{fhwa2023a}, which is shown in Table~\ref{tab:func_class}. Using this coding scheme, our targeted surface street roadways would exclude “Interstates” and “Other Arterials - Other Freeways and Expressways”. Functional system was only available for FARS, ADOT CPM, and FHWA mileage. None of the other mileage and crash datasets use this coding scheme directly.

\begin{table*}[width = 0.6\textwidth, cols=2, pos=h]
  \caption{The highway functional classification coding is presented for each functional system and aggregated at a higher categorical level.}
  \label{tab:func_class}
  \centering
  \begin{tabular*}{\tblwidth}{|c|c|}
    \cline{1-2}
    \textbf{Higher category} & \textbf{Functional System} \\
    \cline{1-2}
    Interstate System & Interstates \\
    \multirow{3}{*}{Other Arterials} & Other Freeways and Expressways \\
    \cline{2-2}
    & Other Principal Arterial \\
    \cline{2-2}
    & Minor Arterials \\
    \cline{1-2}
    \multirow{2}{*}{Collectors} & Major Collectors \\
    \cline{2-2}
    & Minor Collectors \\
    \cline{1-2}
    Local & Local \\
    \cline{1-2}
  \end{tabular*}
\end{table*}

Two concurrent goals were optimized for when defining our surface street classification scheme: (1) exclude non-surface street road types and (2) align the mileage and crash data road type definitions to ensure apples-to-apples merging of the two datasets. Alignment of road type definitions between the mileage and crash data limits risk of over or under estimating mileage exposure, which would directly affect the computed ADS crash rate. Table~\ref{tab:surface_streets} covers the various encodings used to best identify surface streets when forming the various driver datasets. 

\begin{table*}[width = 0.95\textwidth, cols=4, pos=h]
  \caption{The unique set of variables and values leveraged from each driver dataset to uniquely identify surface streets. For the range of values presented, a ``:'' is used to indicate a sequence of values from some start to end value. For example, ``2:4'' implies the values ``2, 3, and 4.''}
  \label{tab:surface_streets}
  \centering 
  \begin{tabular*}{\tblwidth}{|P{0.6in}|P{0.6in}|P{2.5in}|P{1.5in}|}
    \cline{1-4}
    \textbf{Geographic Region} & \textbf{Severity Level} & \textbf{Crash} & \textbf{Mileage} \\
    \cline{1-4}
    \multirow{2}{*}{National} & All police-reported & CRSS: \texttt{int\_hwy} equal to 0 FARS: \texttt{func\_sys} IN (2:7) & Exclude Interstate \\
    \cline{2-4}
    & Fatal Only & \texttt{func\_sys} equals (3:7) & Exclude Interstate and ``Other Freeways and Expressways'' \\
    \cline{1-4}
    Maricopa County & All police-reported & \texttt{GeocodeOnRoad} contains ('Ave', 'Rd', 'Blvd', 'Dr', 'Pl', 'St', 'Ln', 'Way', 'Ct', 'Pkwy', 'Hwy', 'Trl', 'Cir', 'Loop','Calle','Via', 'Mc 85', 'SR-74', 'SR-85', 'SR-87', 'SR-303', 'SR-347', 'SR-88','SR-8B', 'SR-238', 'SR-153', 'SR-587', 'SR-71', 'SR-188') \newline \textbf{OR} \newline PostedSpeed $<=$ 45 mph & Exclude Interstate and ``Other Freeways and Expressways'' \\
    \cline{1-4}
    San Francisco County & All police-reported & \texttt{Chp\_beat\_type} is not equal to (1, 2, 3) & Exclude roadways with state highway jurisdiction \\
    \cline{1-4}
    Los Angeles County & All police-reported & \texttt{Chp\_beat\_type} is not equal to (1, 2, 3) & Exclude roadways with state highway jurisdiction \\
    \cline{1-4}
  \end{tabular*}
\end{table*}

For all national, non-fatal crash rates, we were limited to only excluding interstates due to limitations in available CRSS variables. Specifically, there is a variable to indicate interstates in CRSS, but no variables to robustly identify all other highway variants. Accordingly, we prioritized aligning the mileage and crash data by only removing interstates. This did result in the national, non-fatal dataset including ``Other Arterials: Other Freeways and Expressways.''

For national fatality crash rates, FARS crashes and FHWA mileage directly use functional system coding. In both datasets, we were able to identify and exclude both interstates and ``Other Freeways and Expressways.''

Using ADOT data, we were able to identify surface streets using a combination of road name suffix and speed limit, and we attempted to match the manual coding logic to the mileage data, which had a functional system directly encoded. 

Lastly, using California data, we excluded all crashes on interstates, state routes, and US highways and all mileage on roadways with state highway jurisdiction. This enabled us to roughly match the mileage and crash data by approximately removing all higher speed roads. There were some lower speed limit stretches of state highways being excluded in both the mileage and crash data.

\subsubsection{Passenger vehicles}

In-transit, passenger vehicles were identified in the national and state crash databases. The variables and values relied upon for capturing this passenger vehicle population from the crash databases can be found in Table~\ref{tab:passenger_vehicles}. 

\begin{table*}[width = 5in, cols=3, pos=h]
  \caption{The criteria used to identify passenger vehicles in all crash databases. Both “Passenger Vehicle” and “In-Transport Status” must be satisfied for inclusion.}
  \label{tab:passenger_vehicles}
  \centering
  \begin{tabular*}{\tblwidth}{|P{0.6in}|P{3.5in}|P{1.2in}|}
    \cline{1-3}
    \textbf{Region} & \textbf{Variable Name \& Value} & \textbf{Captures} \\
    \cline{1-3}
    \multirow{3}{*}{\parbox{0.6in}{\centering National}} &  \texttt{BODY\_TYP} in (1:17, 19:25, 28:42, 45:49) & Passenger Vehicle \\
    \cline{2-3}
    & \texttt{BODY\_TYP} in (98, 99) & Vehicle - NFS \\
    \cline{2-3}
    & \texttt{UNITTYPE} equal to 1 & In-Transport Status \\
    \cline{1-3}
    \multirow{4}{*}{\parbox{0.6in}{\centering Maricopa County}} & \texttt{BodyStyle} in (2:7, 9:26, 30:32, 34:53, 71, 72) & Passenger Vehicle \\
    \cline{2-3}
    & \texttt{BodyStyle} in (94:104, 107:109, 112:114, 116:119) & \multirow{2}{*}{Vehicle} \\
    \cline{2-2}
    & \texttt{UnitType} equals 1 \textbf{AND}\newline\texttt{BodyStyle} equals (-1, 254, 255) & \\
    \cline{2-3}
    & \texttt{UnitAction} not\_in (14, 15) & In-Transport Status \\
    \cline{1-3}
    \multirow{4}{*}{\parbox{0.6in}{\centering San Francisco and Los Angeles Counties}} & \texttt{Stwd\_vehicle\_type} in (A, B, D, E) \textbf{OR}\newline\texttt{Chp\_veh\_type\_towing} in (1, 7, 8, 21, 22, 23, 48, 71:73, 81:83) & Passenger Vehicle \\
    \cline{2-3}
    & \texttt{Stwd\_vehicle\_type} in (J, M) \textbf{OR}\newline\texttt{Chp\_veh\_type\_towing} in (32, 34:36, 98) & \multirow{2}{*}{Vehicle - NFS} \\
    \cline{2-2}
    & \texttt{party\_type} equals 1 \textbf{AND} \newline \texttt{Stwd\_vehicle\_type} equals NULL \textbf{AND} \newline \texttt{Chp\_veh\_type\_towing} equals NULL & \\
    \cline{2-3}
    & \texttt{move\_pre\_acc} NOT equal to 0 \textbf{AND} \newline Party\_type NOT equal to 3 & In-Transport Status \\
    \cline{1-3}
  \end{tabular*}
  
  \footnotesize{$^1$ The same variable names and values were used for both CRSS and FARS databases.}
\end{table*}

We targeted vehicles with a gross vehicle weight rating (GVWR) of 10,000 pounds or less, which is consistent with 49 CFR § 565.15 classification of passenger vehicles \citep{vinrequirements2022}. This definition also matches the class-1 and class-2 GVWR classifications readily identifiable from vehicle identification numbers (VIN) and the Class 2 and 3 vehicle types in FHWA highway statistics annual reporting \citep{ncsa2022vpic, fhwa2023a}. 

A proportion of the crashed vehicles had enough information to identify the actor as a vehicle, but not enough information to indicate whether or not the vehicle was a passenger vehicle. For these vehicles not further specified (NFS), we imputed vehicle type by applying a weighting factor that represented the total proportion of known vehicles that were passenger vehicles. The computed weighting factors applied to NFS vehicles for national, Maricopa county, San Francisco county, and Los Angeles county were 0.94, 0.90, 0.89, and 0.93, respectively. 

In addition, only vehicles “in-transit” were included in this study. Vehicles not “in-transit” were readily identifiable in all datasets and would include, among others, vehicles parked in designated parking, vehicles parked off the roadway, vehicles parked on private property, and working vehicles. 

All mileage data sources provided mileage totals across all vehicle types, which includes single trucks, combination trucks, buses, and motorcycles. We relied on the FHWA VM-4 tables to identify which mileage was attributable to passenger vehicles, which are labeled as ``passenger cars'' and ``light trucks''. The VM-4 tables provide independent proportions by both functional systems (broken down by interstate, other arterial, and other), census-defined urban/rural population grouping, and U.S. state \citep{ohpi2016}. 

For all national mileage estimates, combining the VM-2 and VM-4 tables was straightforward. Both datasets are broken down by functional system, urban/rural designation, and state. The VM-4 proportion of passenger vehicles for these groupings was simply applied to the appropriate cell of the VM-2 table to generate a passenger vehicle-only version of the VM-2 table. 

When applying the VM-4 table to both California PRD and Arizona CPM mileage data, the urban VM-4 table was exclusively used for the corresponding U.S. state. Similar to the national crash data, the Arizona CPM mileage data is broken down by functional systems, which enables the proportion of passenger vehicles to be directly applied to the overall mileage estimates. For the California PRD data surface streets, the FHWA VM-4 table ``other arterial'' and ``other'' passenger vehicle percentages were averaged together to generate a single percentage of passenger cars on surface streets. This percentage was applied to the total surface street mileage to generate the passenger car mileage on surface streets. 

\subsubsection{Severity level}
\label{appendix:severity_level}

\textit{Any-injury-reported} crash rates, where an Injury occurred, were identified using the ``KABCO'' injury classification scale for all driver datasets \citep{ghsa2017mmucc}. Specifically, only crashes with a reported injury level (``K'', ``A'', ``B'', or ``C'') were included to generate these crash rates. For the national crashes identified within CRSS, we relied upon the ``MAXSEV\_IM'' variable, which uses the KABCO reported level but incorporates imputation in the event of unknown or missing data. In addition, CRSS crashes with ``Injured, Severity Unknown'' (MAXSEV\_IM equal to 5) were included. The non-fatal injury crash underreporting adjustment by \cite{blincoe2023economic} was applied to the non-fatal injury crashes. 

A \textit{tow-away} crash is one in which any vehicle involved was towed away from the crash scene. \textit{Tow-away} crashes are not mutually exclusive from \textit{any-injury-reported} crashes, so there is overlap in some crashes that contribute to both the \textit{tow-away} crash rate and \textit{any-injury-reported} crash rate. Tow-away crashes were readily identifiable in both state crash databases at the crash-level (ADOT: ``TowAwayFlag''; SWITRS: ``tow\_away''). For national rates, the ``towed'' variable (equal to 2, 3, 6, or 7) present in both CRSS and FARS for every vehicle was used. If any vehicle within a given crash was towed, the crash was indicated to be a tow-away. 

\textit{Any airbag deployment} crashes were those in which any involved vehicle had an airbag deployment. In CRSS and FARS, the person table was used to identify airbag deployments (``air\_bag'' equal to 1, 2, 3, 7, 8, or 9). Using CA SWITRS data, both the ``Party'' (person-level) and ``Unit'' (vehicle-level) tables were used, where an indicated deployment by either a vehicle or an involved person would classify the event as an airbag deployment crash. If any person (``victim\_safety\_equip\_1'' or ``victim\_safety\_equip\_2'' equal to ``L'' or ``M'') or vehicle (``party\_safety\_equip\_1'' or ``party\_safety\_equip\_2'' equal to ``L'' or ``M'') indicated a deployment, the crash was labeled as an airbag deployment. Using ADOT data, the ``Person'' tables was used. If person-level ``Airbag'' (equal to 2, 3, 4, 5, 102, 103, or 105) or ``SafetyDevice'' (equal to 6 or 7) indicated a deployment, the crash was labeled accordingly.  

A \textit{suspected serious injury+} crash was one in which any involved person sustained either a suspected serious injury or fatal injury (KABCO is either ``K'' or ``A''). \textit{Fatal} crashes were those in which any person involved was fatally injured as a result of the crash events. \textit{Fatal} crashes were identified from state databases using the ``K'' classification in the KABCO crash severity score. All crashes in FARS are, by inclusion, \textit{fatal} crashes.

\subsection{Underreporting Adjustment Factors}
\label{appendix:underreport_adjust_fact}

Table~\ref{tab:correction_factors} shows the crash-level underreporting correction factors relied upon in the current study, which were applied to each police-reported crash using an adjustment correction factor.

\begin{table*}[width = 3.5in, cols=4, pos=h]
  \caption{A complete listing of correction factors used in the current study to account for underreporting in police-reported data.}
  \label{tab:correction_factors}
  \centering
  \begin{tabular*}{\tblwidth}{|c|c|c|c|}
    \cline{1-4}
    \multirow{2}{*}{\textbf{Crash Severity}} & \multirow{2}{*}{\textbf{Percent Underreported}} & \multicolumn{2}{c|}{\textbf{Correction Factor}} \\
    \cline{3-4}
    & & \textbf{Blincoe} & \textbf{SHRP-2} \\
    \cline{1-4}
    PDO & 59.7\% & 2.48 & 6.25 \\
    \cline{1-4}
    Non-Fatal Injury & 31.9\% & \multicolumn{2}{c|}{1.47} \\
    \cline{1-4}
    Fatal & 0.0\% & \multicolumn{2}{c|}{1.00} \\
    \cline{1-4}
  \end{tabular*}
\end{table*}

\subsection{Errors When Computing a Vehicle-level Crash Rate}
\label{appendix:veh_level_crash_rate}

We identified multiple publications that improperly compared human crash-level rates to ADS vehicle-level rates \citep{banerjee2018hands, blanco2016automated, cummings2024assessing, favaro2017examining, kalra2016driving, schoettle2015preliminary}. Crash rates derived from ADS data represent the number of times an ADS crashes for some amount of VMT (a vehicle-level or driver-level crash rate). The appropriate benchmark comparison to this ADS rate would also be a vehicle-level crash rate, or the number of times human drivers crashed over some amount of VMT. 

In the undercounting of the human driver crash rate, rather than determine the number of human drivers who crashed (a vehicle-level or driver-level rate), these past studies determined the number of overall crashes (a crash-level rate). The total number of crashes is not equivalent to the total number of drivers that crashed, because crashes, more often than not, involve multiple vehicle drivers. Using crash counts instead of the number of crashed vehicles leads to undercounting of crashed vehicle rates, which specifically underestimates the human crash rate (a better safety performance than in reality) and does not represent the number of human drivers who crashed divided by the mileage they had accumulated.

In a review of multiple prior benchmarking studies, \citet{young2021critical} claimed that a vehicle-level crash rate is improper but failed to provide any physical or mathematical reasoning.  This was the only acknowledgement of incongruence in crash rate calculations (crash-level vs. vehicle-level) between studies that the authors could find when examining the literature. Young's (2021) assertion, however, is incorrect. As an exercise, consider a town of 6 vehicles (A, B, C, D, E, and F) that each drive 2,000 miles in a year for a total of 12,000 VMT in the town. Over the course of that year, two of the vehicles, A and B, get into a crash with one another. In a separate incident, vehicle C also runs off the road, and has a crash with a fixed object. There are no other crashes. We want to know: how many times do drivers in this town crash per VMT? In the incorrect computation, the crash count would be “2”, and the following conclusion would be drawn: drivers crash once every 6,000 VMT. This formulation discounts the fact that there were two crashes partners in the crash between vehicle A and B. Imagine that each of these vehicles was equipped with sensors that can detect crashes (like an event data recorder or a perception system) and all mileage is being tracked onboard. At the vehicle-level, we would have observed “3” crashed vehicles over the course of 12,000 miles, and this would draw the conclusion: drivers crash once every 4,000 VMT. This is exactly the way vehicle crash rates are computed in NDS studies and with ADS fleets: the number of times the driver(s) crashed over some amount of VMT. This fictitious example shows how easy it is, on one hand, to compute very different crash rates out of the same dataset and, on the other, how nuanced the process for computing appropriate performance crash rates is. 

We estimate what the effects of this commonly performed error would be using the results from the current study. Using the 2022 national police-reported crashes and VMT in Table~\ref{tab:benchmark_results}, there were 5.9 million police-reported crashes in 2021 over 3.2 trillion VMT. These 5.9 million crashes involved a total of 10.5 million crashed vehicles, which computes to 3.28 police-reported crashed vehicles per million miles (an average of 1.78 drivers in every crash). The result is that the vehicle-level rate is 78\% higher than the crash-level rate using the national crash statistics. So, making this mathematical error when comparing to ADS vehicle-level rates would serve to greatly underestimate ADS performance in a comparison.  

Another important quoted rate is the number of fatalities (fatally injured persons) per VMT. In 2021, there were 42,939 people fatally injured in crashes on U.S. public roadways, which equates to a fatality rate of 1.37 persons per 100 million VMT \citep{stewart2023overview}. This fatality rate represents the number of persons killed over an entire population’s VMT, which is inherently different from a crashed vehicle rate - the number of times a vehicle was involved in a fatal crash divided by the number of miles driven. In \citet{kalra2016driving} influential study computing the amount of driving miles needed for “demonstrating autonomous vehicle safety,” the fatality rate was incorrectly used to represent the vehicle crash rate. In 2021, there were 61,332 vehicles involved in fatal crashes in the U.S., which computes to 1.96 vehicles (or drivers) involved in fatal crashes per 100 million VMT (crashed vehicle rate). The U.S. vehicle fatal crash rate was 43\% higher than the fatality rate in 2021.  

Few of the previously published studies correctly compared an ADS crashed-vehicle rate to a benchmark crashed-vehicle rate. A study done by \citet{teoh2017rage} compiled the police-reported crash data at the driver-level for comparison to the ADS fleet crash rates. In their study, \citet{teoh2017rage} present the benchmark data in “human-driven passenger vehicles involved in police-reported crashes per million VMT.” Similarly, \citet{lindman2017basic} used the Swedish Transport Accident Data Acquisition (STRADA) police-reported database to estimate the required driving mileage for measuring statistically significant differences in crash rates between automated vehicles and a benchmark. The benchmark from \citet{lindman2017basic} also correctly considered the crash data at the vehicle-level. 

\subsection{Underreporting of Reported Crashes to SWITRS}
\label{appendix:switrs_underreporting}

Systematic underreporting of police-reported PDO crashes to SWITRS (California’s state crash database) adds an additional layer of surveillance bias not addressed by this study’s underreporting corrections. To reiterate this limitation, the SWITRS database, by design, is intended to capture all reported injuries and fatalities. With regards to PDO-level crash reports, “some agencies report only partial numbers of their PDO crashes, or none at all” \citep{chp2021}. This is particularly troubling given that this would artificially deflate both SGO reportable and police-reported estimates - another underestimation-inducing bias brought on by the available human crash data. 

From the available literature and data tables, we were unable to identify if this PDO reporting to SWITRS applied to San Francisco and Los Angeles counties. We did, however, note that the proportion of PDO (no injuries) to \textit{any-injury-reported} crashes varied between San Francisco and Los Angeles despite being located in the same state and having the same nominal reporting threshold. When looking at passenger vehicles on surface streets, approximately one-half (52\%) of SWITRS reported crashes were PDO in Los Angeles, whereas PDO crashes in San Francisco were only one-third (33\%) of the total PDO reported events. For comparison, over two-thirds of Maricopa County (72\%) and national (71\%) police-reported crashes were PDO. Arizona is likely influenced by the lower PDO reporting threshold, and there are a range of reporting thresholds used nationally. It is not clear whether these differences are due to (a) systematic reporting differences or (b) heterogeneity in the proportion of \textit{any-injury-reported} to PDO crashes. The explicit SWITRS requirement that it “processes all reported fatal and injury crashes that occurred on California’s state highways and all other roadways, excluding private property” indicates that this data is most useful when examining higher severity crashes, and not intended for PDO comparisons \citep{chp2021}.

\subsection{Year-to-Year Effects on Crash Rates}
\label{appendix:year_effects_crash_rates}

Crash rates vary from year-to-year \citep{national2023early}. This appendix section presents how the study's generated benchmarks have varied over the previous years of available data. There are many factors that affect changes in crash rates. Rather than explore these factors with respect to year, the purpose of showing these results is simply to highlight the importance of considering crash and mileage data years in any analysis.

\begin{figure}[h!]
    \centering
    \includegraphics[width=6in]{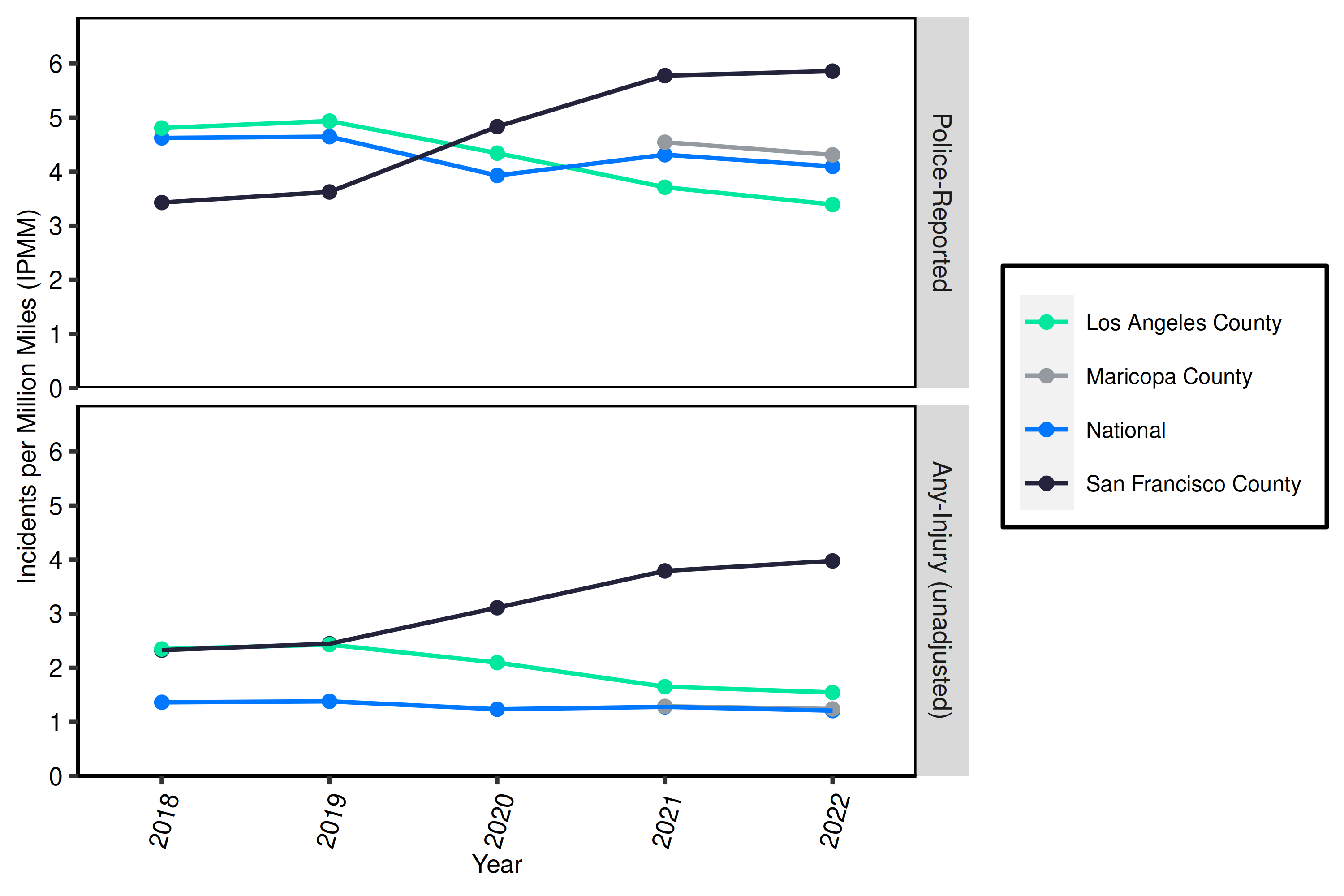}
    \caption{Year-to-year variations in \textit{police-reported} and \textit{unadjusted any-injury-reported} ADS-comparable benchmarks in currently deployed geographic regions.}
    \label{fig:year_effects_benchmarks}
\end{figure}

To demonstrate this yearly dependency, \textit{police-reported} and \textit{unadjusted any-injury-reported} crash rates are presented in Figure \ref{fig:year_effects_benchmarks}. National and California crash rates are presented over the last five years of available data (2018 through 2022). Maricopa County crash rates were restricted to only 2021 and 2022 due to a lack of available driving mileage exposure data for before 2021 on the ADOT "Extent and Travel" Dashboard 
\citep{arizona2023b}. There are two important findings. First, there was year-to-year variation in all analyzed geographic areas. Second, the direction and magnitude of this year-to-year trending has geographic and severity level dependency. The national crash data notably dipped in 2020 with a larger percentage drop in \textit{police-reported} crash rates when compared to \textit{unadjusted any-injury-reported} rates. Los Angeles and San Francisco county data both had notable trends beginning in 2020 for all severity levels with each county's crash rates trending in opposite direction. For the two years of available Maricopa county data, the magnitude and direction of the crash rates were nearly identical to the national estimates. 

The trends starting in 2020 are likely highly influenced by the COVID-19 pandemic, which was affecting travel behaviours in the US. It is unclear that crash data from before the pandemic is representative of the post-pandemic crash population. Regardless, the importance of controlling for year-to-year variations is apparent. 

A comprehensive listing of crash rates across all geographic areas and severity levels is presented for 2021 data in Table \ref{tab:benchmark_2021_results}. The results provide a reference for researchers comparing to 2021 data. 

\newgeometry{margin=1cm} %
\begin{landscape}
\renewcommand{\arraystretch}{1.35}

\begin{table*}[width=8.5in, cols=9, pos=h]
  \caption{2021 Data: A full compilation of the ADS-relevant crash rate benchmarks (presented in crashed vehicle incidents per million/billion Miles, IPMM / IPBM) with intermediate values is presented. Mileage is presented in million miles (Mmi). This includes crash rate tabulations by geographic region and after adjusting for vehicle type and road type.}
  \centering
    \begin{tabular*}{\tblwidth}{|c|r|r|c|c|c|c|c|c|}
      \cline{5-8}
      \multicolumn{4}{c|}{} & \textbf{National} & \textbf{Maricopa County} & \textbf{San Francisco County}	& \textbf{Los Angeles County} \\
      \cline{1-9}
      \multicolumn{2}{|c|}{\multirow{5}{*}{\rotatebox{90}{\parbox{1in}{\centering\textbf{All Police-reported}}}}} & \multicolumn{2}{r|}{\textbf{Mileage}} & 3,132,411 Mmi & 41,291 Mmi & 2,238 Mmi & 69,643 Mmi & \multirow{8}{*}{\rotatebox{270}{\centering\textbf{Intermediate Values}}} \\
      \cline{3-8}
      \multicolumn{2}{|c|}{} & \multirow{2}{*}{\textbf{Crashes}} & \multirow{2}{*}{Unadjusted} & \multirow{2}{*}{6,102,936} & \multirow{2}{*}{86,747} & \multirow{2}{*}{5,776} & \multirow{2}{*}{108,708} & \\
      \multicolumn{2}{|c|}{} & & & & & & & \\
      \cline{3-8}
      \multicolumn{2}{|c|}{} & \multirow{2}{*}{\parbox{1.5in}{\flushright\textbf{All Vehicles, Any Type (IPMM)}}} & \multirow{2}{*}{Unadjusted} & 10,843,740 & 168,985 & 9,312 & 196,466 & \\
      \multicolumn{2}{|c|}{} & & & (3.46 IPMM) & (4.09 IPMM) & (4.16 IPMM) & (2.82 IPMM) & \\
      \cline{1-8}
      \multirow{18}{*}{\rotatebox{90}{\centering\textbf{Crashed Passenger Vehicles}}} & \multirow{3}{*}{\rotatebox{90}{\parbox{0.35in}{\centering\textbf{All Roads}}}} & \multicolumn{2}{r|}{\textbf{Mileage}} & 2,769,006 Mmi & 37,537 Mmi & 1,934 Mmi & 60,180 Mmi & \\
      \cline{3-8}
      & & \multirow{2}{*}{\textbf{Police-Reported (IPMM)}} & \multirow{2}{*}{Unadjusted} & 10,066,367 & 155,390 & 8,306 & 182,140 & \\
      & & & & (3.64 IPMM) & (4.14 IPMM) & (4.29 IPMM) & (3.03 IPMM) & \\
      \cline{2-9}
      & \multirow{15}{*}{\rotatebox{90}{\centering\textbf{Surface Streets}}} & \multicolumn{2}{r|}{\textbf{Mileage}} & 2,109,149 Mmi & 24,224 Mmi & 927 Mmi & 28,445 Mmi & \multirow{15}{*}{\rotatebox{270}{\textbf{ADS Benchmarks}}} \\
      \cline{3-8}
      & & \multirow{4}{*}{\textbf{\parbox{1.5in}{\flushright Any Property Damage or Injury (IPMM)}}} & \multirow{2}{*}{Blincoe-Adjusted} & 19,822,830 & 241,568 & 9,703 & 214,233 & \\
      & & & & (9.40 IPMM) & (9.97 IPMM) & (10.5 IPMM) & (7.53 IPMM) & \\
      \cline{4-8}
      & & & \multirow{2}{*}{Blanco-Adjusted} & 43,946,719 & 539,276 & 16,629 & 435,328 & \\
      & & & & (20.8 IPMM) & (22.2 IPMM) & (18.0 IPMM) & (15.3 IPMM) & \\
      \cline{3-8}
      & & \multirow{2}{*}{\textbf{Police-Reported (IPMM)}} & \multirow{2}{*}{Unadjusted} & 9,096,907 & 110,207 & 5,350 & 105,625 & \\
      & & & & (4.31 IPMM) & (4.55 IPMM) & (5.77 IPMM) & (3.71 IPMM) & \\
      \cline{3-8}
      & & \multirow{4}{*}{\parbox{1.5in}{\flushright\textbf{Any-Injury-Reported (IPMM)}}} & \multirow{2}{*}{Unadjusted} & 2,695,591 & 31,210 & 3,512 & 46,958 & \\
      & & & & (1.28 IPMM) & (1.29 IPMM) & (3.79 IPMM) & (1.65 IPMM) & \\
      \cline{4-8}
      & & & \multirow{2}{*}{Blincoe-Adjusted} & 3,938,741 & 45,546 & 5,143 & 68,657 & \\
      & & & & (1.87 IPMM) & (1.88 IPMM) & (5.55 IPMM) & (2.41 IPMM) & \\
      \cline{3-8}
      & & \multirow{2}{*}{\textbf{Tow-Away (IPMM)}} & \multirow{2}{*}{Unadjusted} & 4,029,840 & 50,007 & 2,425 & 50,927 & \\
      & & & & (1.91 IPMM) & (2.06 IPMM) & (2.62 IPMM) & (1.79 IPMM) & \\
      \cline{3-8}
      & & \multirow{2}{*}{\textbf{Any Fatality}} & \multirow{2}{*}{Unadjusted} & 39,354 & 605 & 31 & 635 & \\
      & & & & (18.7 IPBM) & (25.0 IPBM) & (33.2 IPBM) & (22.3 IPBM) & \\
      \cline{1-9}
      
    \end{tabular*}
    \label{tab:benchmark_2021_results}
\end{table*}

\end{landscape}
\restoregeometry

\renewcommand{\arraystretch}{1.0}

\subsection{Power Analysis Comparison to Kalra and Paddock}
\label{appendix:kalra_paddock}

The statistical power analysis performed in this paper is not unlike the analysis shown in Figure 3 of \citet{kalra2016driving}. As discussed in Appendix \ref{appendix:veh_level_crash_rate}, \citet{kalra2016driving} improperly used a fatality rate (fatally injured persons per VMT) rather than a vehicle-level crash rate. Additionally, \citet{kalra2016driving} used an earlier year of crash data (2013), and generated benchmark rates that included all vehicle types and driving conditions including freeways, whereas the national estimates from this study are generated for passenger vehicles only and exclude interstate highways. These differences resulted in the national benchmark rates from this study to be slightly larger than those used by \citet{kalra2016driving} (22.5 IPBM vs 10.9 IPBM for fatal, 1.78 IPMM vs 1.03 IPMM for \textit{Blincoe-adjusted any-injury-reported}, and 9.44 IPMM vs 3.82 IPMM for \textit{Blincoe-adjusted any property damage or injury}).Therefore, the analysis presented in this paper and from \citet{kalra2016driving} find approximately similar necessary ADS miles to determine statistical significance.

The conclusion of \citet{kalra2016driving} was generally that it is infeasible to use public road testing data to draw conclusions about ADS safety performance due to the large number of miles needed. The
analysis of \citet{kalra2016driving} assumed modest differences between the ADS and the benchmark crash
rates, using an ADS that has a crash rate of 80\% of the benchmark as a reference point. This may have been
a prudent reference at the time when only early testing data from ADS was available. The results of early ADS safety impact analysis using RO data \citep{kusano2023comparison} suggest that the \textit{police-reported} and \textit{any-injury-reported} performance of an ADS fleet operating in RO operations generally have lower crashed vehicle rates compared to the benchmarks. \textit{Police-reported} crashes reduced between 51\% and 70\% (i.e., a \textit{police-reported} crash rate that is between 49\% and 30\% of the human benchmark). The larger the reduction in crash rates due to the ADS, the fewer miles are needed for statistical significance. For example, for the \textit{police-reported} benchmark, Table~\ref{table:power} showed that an ADS with 25\% the human benchmark would require 5.4 million miles compared to 61.9 million miles for an ADS that had 75\% the human benchmark, an order of magnitude difference in the required miles.

Another difference between the statistical power analysis done in this study and Figure 3 of \citet{kalra2016driving} was that this paper rounded the number of expected events to the next lowest integer, whereas \citet{kalra2016driving} appeared to have computed the number of miles using fractional event counts. This results in a smooth and monotonically curve with the number of miles needed for significance as a function of ADS performance relative to the benchmark.

\end{document}